\documentclass[10pt,twocolumn,letterpaper]{article}

\usepackage{iccv}              % To produce the CAMERA-READY version
\definecolor{iccvblue}{rgb}{0.21,0.49,0.74}
\usepackage[pagebackref,breaklinks,colorlinks,allcolors=iccvblue]{hyperref}

\def\paperID{408} % *** Enter the Paper ID here
\def\confName{ICCV}
\def\confYear{2025}

\title{FreeDNA: Endowing Domain Adaptation of Diffusion-Based Dense Prediction with Training-\underline{Free} \underline{D}omain \underline{N}oise \underline{A}lignment}
\usepackage{graphicx}
\usepackage{algorithm}
\usepackage{algorithmic}
\usepackage{multirow}
\author{Hang Xu$^{1}$ \quad Jie Huang$^{1}$ \quad Linjiang Huang$^{2}$ \quad Dong Li$^{1}$ \quad Yidi Liu$^{1}$ \quad Feng Zhao$^{1}$\thanks{Corresponding author}\\
$^{1}$University of Science and Technology of China $^{2}$Beihang University\\
}

\begin{document}
\maketitle

\begin{abstract}
Domain Adaptation(DA) for dense prediction tasks is an important topic, which enhances the dense prediction model's performance when tested on its unseen domain.
% which adapts a model trained on the source domain to perform better on the target domain. 
Recently, with the development of Diffusion-based Dense Prediction (DDP) models,  the exploration of DA designs tailored to this framework is worth exploring, since the diffusion model is effective in modeling the distribution transformation that comprises domain information. 
In this work, we propose a training-free mechanism for DDP frameworks, endowing them with DA capabilities. 
Our motivation arises from the observation that the exposure bias (e.g., noise statistics bias) in diffusion brings domain shift, and different domains in conditions of DDP models can also be effectively captured by the noise prediction statistics.
% and mitigating it through adjusting noise statistics leads to correcting the domain shift. 
% Additionally, we further discover that in diffusion-based dense prediction models, the differences in the generation process of conditions from different domains can also be effectively captured by the statistics of noise prediction. 
Based on this, we propose a training-free Domain Noise Alignment (DNA) approach, which alleviates the variations of noise statistics to domain changes during the diffusion sampling process, thereby achieving domain adaptation. 
Specifically, when the source domain is available, we directly adopt the DNA method to achieve domain adaptation by aligning the noise statistics of the target domain with those of the source domain.
For the more challenging source-free DA, inspired by the observation that regions closer to the source domain exhibit higher confidence meeting variations of sampling noise, 
we utilize the statistics from the high-confidence regions progressively to guide the noise statistic adjustment during the sampling process.
% we further introduce a noise adjustment estimation mechanism to assist the strategy.
% We utilize the statistics from high-confidence regions to guide the noise adjustment during the sampling process, followed by further refinement for the confidence estimation and guidance mechanism. 
Notably, our method demonstrates the effectiveness of enhancing the DA capability of DDP models across four common dense prediction tasks. Code is available at \href{https://github.com/xuhang07/FreeDNA}{https://github.com/xuhang07/FreeDNA}.
\end{abstract}    
\begin{figure}[t]
	\centering
	\includegraphics[width=0.95\linewidth]{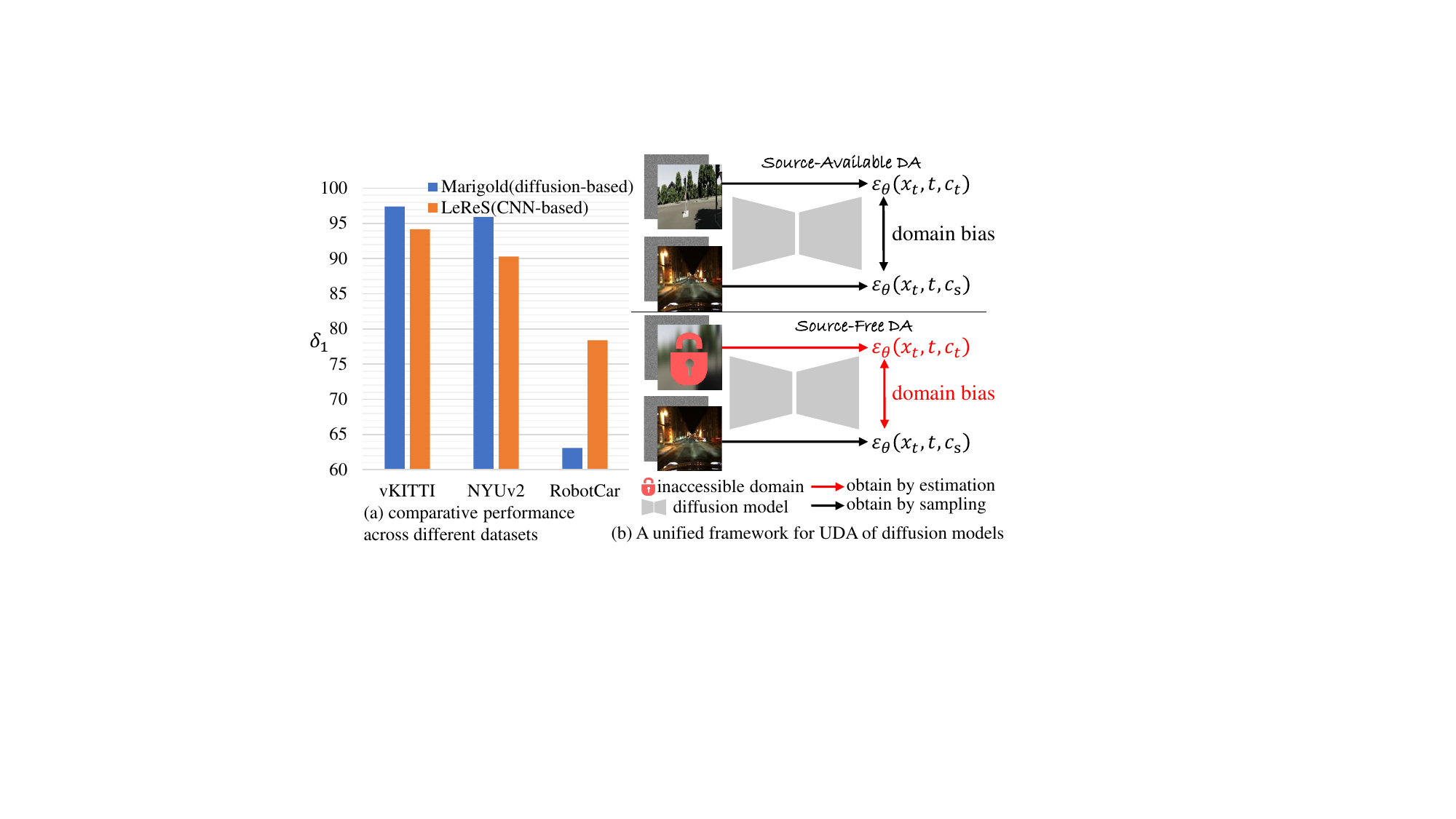}
	\caption{Take depth estimation as an example. In (a), the trained Marigold~\cite{ke2024repurposing} is directly tested on datasets with different domains, whereas LeReS~\cite{LeReS} is respectively trained and tested on these datasets. Marigold depicts robust zero-shot capabilities on datasets close to source domain it is trained (vkitti dataset) but drops in performance compared with LeReS on datasets farther from source domain, highlighting the necessity of domain adaptation for diffusion-based dense prediction (DDP). (b) shows our proposed framework for DDP's domain adaptation, aiming to improve performance on target domain by alleviating domain bias.} 
	\label{fig:first}
	\vspace{-10pt}
\end{figure}

\section{Introduction}
\label{sec:intro}
Deep learning models have been proven effective on dense prediction tasks~\cite{ranftl2021vision,yuan2021hrformer,wang2021pyramid}, while they may struggle when encountering unseen domain data~\cite{london2016stability,tateno2018distortion}. To address this, domain adaptation (DA) has been proposed to adapt the models trained on the source domain to generalize better on the unlabeled target domain, avoiding expensive data annotation and model retraining~\cite{ben2006analysis,wang2018deep}. 

% Diffusion models have already been incorporated into many DA methods, but their application has been limited to data augmentation or source domain data generation.
On the other hand, the Diffusion-based Dense Prediction (DDP) design has demonstrated exceptional performance in many dense prediction tasks~\cite{ji2023ddp,lee2024exploiting},
such as depth-estimation~\cite{he2024lotus,ke2024repurposing}, optical flow estimation~\cite{luo2024flowdiffuser,saxena2023surprising},\textit{etc.}.
% , and \textit{etc.}.
% image super-resolution~\cite{} and image segmentation~\cite{}.
Similarly, these methods also face challenges in generalization when testing them on unseen domains, as shown in~Fig.~\ref{fig:first}.
Since the diffusion model has proven to be effective in modeling distribution transformation in a zero-shot manner~\cite{li2023your,clark2023text}, which often comprises domain information such as style~\cite{hemati2023cross,niemeijer2024generalization}, we believe that developing the DA method tailored to the above DDP models is worth exploring. 
Therefore, we aim to design a general DA framework for DDP models, leveraging the zero-shot capabilities in the diffusion model to avoid the high costs of training.

% However, despite the fact that diffusion models also face challenges in generalization, as illustrated in~Fig.~\ref{fig:first}(a), DA methods specifically tailored to this framework have yet to be explored. Therefore, we aim to design a DA framework specifically tailored for diffusion models, effectively leveraging their inherent zero-shot capabilities to avoid the high costs associated with training.

Our motivation stems from the observation of exposure bias~\cite{ning2023elucidating,ning2023input,li2023alleviating} (e.g., noise statistics bias)
in diffusion models exhibit similar visual changes like domain shift, as shown in~Fig.~\ref{fig:eb_distance}(a).
Since the domain information is mainly relevant to the amplitude component in an image~\cite{chen2021amplitude,kim2023domain}, we further validate that exposure bias is highly related to domain shift by visualizing the amplitude differences of noise prediction between training and testing across various steps in~Fig.~\ref{fig:eb_distance}(b), which illustrates the domain shift brought by exposure bias.
This phenomenon is also confirmed in the DDP model when the conditioned image is the domain shifted (see ~Fig.~\ref{fig:eb_distance}(c)). 
Apart from the above visual analysis, domain bias and exposure bias share several commonalities: their existence stems from the gap between unreliable predictions and reliable predictions~\cite{ning2023elucidating}. 
Therefore, we can refer to the operation in addressing exposure bias (i.e., adjusting the noise prediction statistics) as a potential solution to address the domain bias in DDP~\cite{ning2023elucidating,ning2023input,li2023alleviating}.
We then verify the domain differences in condition images of DDP are clearly illustrated in the DDP's noise prediction statistics in~Fig.~\ref{fig:fourier}, implying that adjusting the noise prediction statistics leads to altering the domain information effectiveness brought by the domain-biased condition image.

% Consequently, in diffusion-based dense prediction models, when the conditions belong to different domains, does such a bias also exist between the predicted noises? The answer is affirmative.
% We refer to the difference in noise prediction as domain bias. Domain bias and exposure bias share several commonalities: their existence stems from the gap between the unreliable predictions and the reliable predictions of the network. 
% Therefore, to mitigate domain bias, it is also necessary to shift the network's predictions from an unreliable vector field to a more reliable one, much like the methods previously employed to address exposure bias. 

% However, there are distinctions between the two: exposure bias does not involve a genuine domain gap. Thus, to resolve domain bias, one must identify a medium that can reflect domain information. Fortunately, noise prediction itself serves as an excellent medium. Through the analysis of the amplitude spectrum in~Fig.~\ref{fig:fourier}, we have discovered a high degree of consistency in stylistic information between noise prediction and output images, which can be simply reflected by first and second-order statistics, inspiring us to adjust the noise prediction according to the variations of it.

\begin{figure*}[t]
	\centering
	\includegraphics[width=1\linewidth]{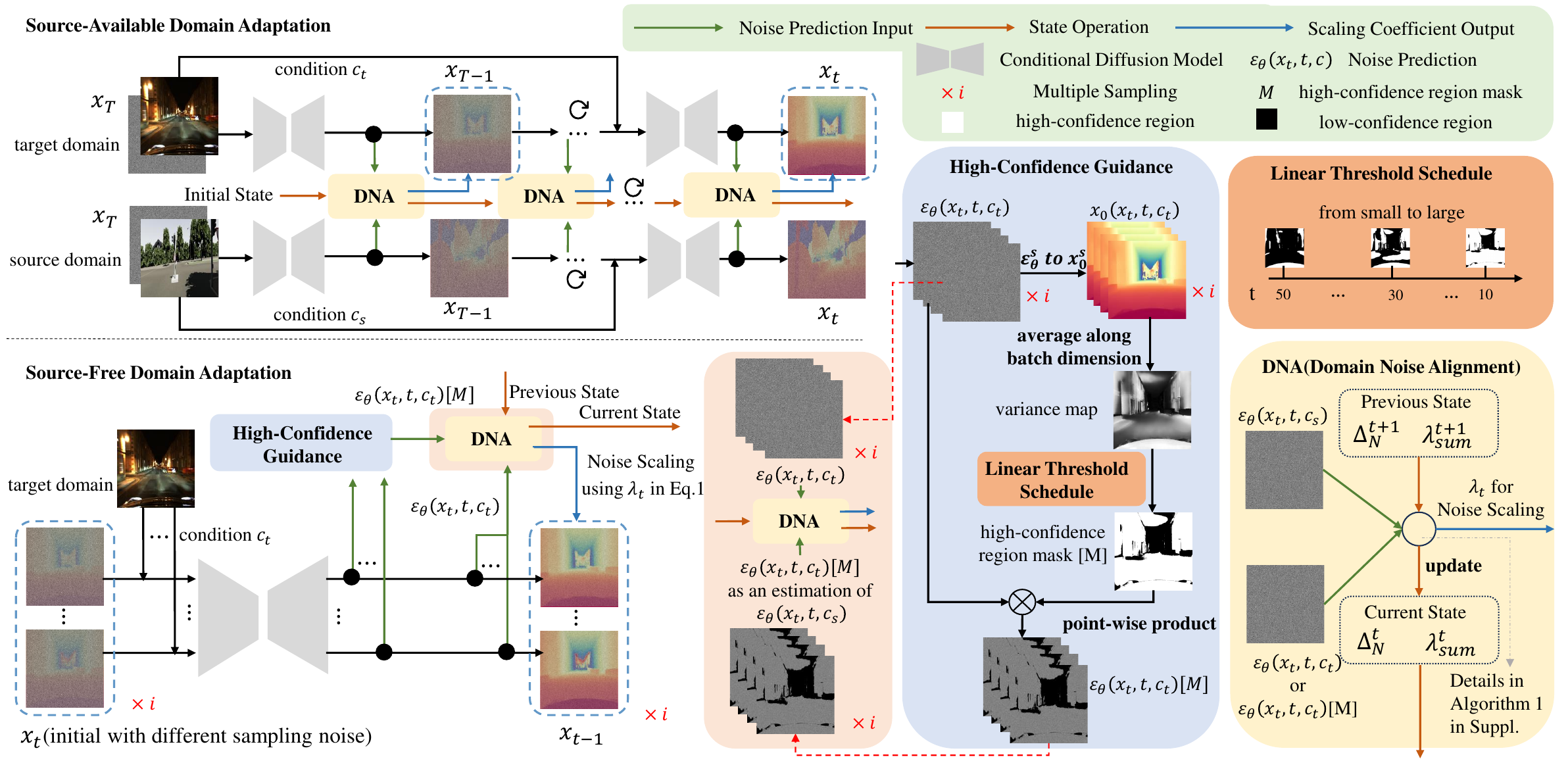}
	\caption{ \textbf{Overall schema of our proposed method.} Our core idea is to adjust the noise prediction of the target domain to align it with the noise prediction of the source domain. When the source domain is known, we directly align the noise with our proposed Domain Noise Alignment. For Source-Free DA, we obtain the high-confidence region mask with the variance map and guide the direction of noise adjustment. DNA takes source domain noise (or estimation) and target domain noise as inputs(\textcolor{green}{green lines}), while also accepting the state from the previous step, updating it, and outputting it to the next DNA block(\textcolor{orange}{orange lines}). Finally, it outputs scaling coefficients $\lambda_t$ to correct the noise estimation(\textcolor{blue}{blue lines}). The pseudocode of DNA for both different settings is provided in the \textbf{supplementary materials}. $\Delta_N$ represents the ratio of the L2 norms of noise from different domains. $\lambda_{sum}$ represents the sum of $\lambda_t-1$ from previous steps.
		 }
	\label{fig:model}
	\vspace{-10pt}
\end{figure*}

% To alleviate the domain bias, 
In this work, we propose a training-free Domain Noise Alignment (DNA) strategy for DDP models by adjusting noise prediction statistics (see Fig.~\ref{fig:model}). 
Specifically, the DNA strategy calculates the variations in noise statistics according to domain changes and leverages them to adjust the diffusion sampling process, thereby achieving domain adaptation. 
For the available source domain, we can directly align the noise prediction to pre-calculated statistics from the source domain with our DNA strategy (see Sec.~\ref{subsec:sada}). For the more challenging source-free DA, we further introduce a noise adjustment estimation mechanism to assist the DNA (see Sec.~\ref{subsec:sfda}). 
Our approach stems from the observation that condition images from the source domain exhibit lower variations with different sampling noise, as shown in~Fig.~\ref{fig:consistency}.  
Based on this, we sample multiple initial noises with the same image as conditions to obtain multiple noise predictions and calculate the variance of noise predictions along the batch dimension, where regions with smaller variances are regarded as high-confidence areas. 
We utilize the statistics from high-confidence regions to guide the noise adjustment during the sampling process. Based on the variations in the sampling process, we further refine the confidence estimation and guidance mechanism across steps to better guide the direction of noise adjustment. 
We summarize our main contributions as follows:

\begin{itemize}[leftmargin=*]
    \item[$\bullet$] We conduct a detailed analysis of the domain bias within diffusion models and delve into the connections and distinctions between it and exposure bias.

    \item[$\bullet$] We propose a training-free domain adaptation method, i.e., Domain Noise Alignment (DNA) for Diffusion-based Dense Prediction (DDP). This method addresses the domain bias by adjusting the noise prediction according to its variation. A noise adjustment estimation mechanism is further introduced for source-free DA.

    \item[$\bullet$] Extensive experiments conducted on four common dense prediction tasks depicts the effectiveness of our method.
\end{itemize}

\section{Related Work}
\label{sec:related}

\subsection{Domain Adaptation for Dense Prediction Tasks}
\vspace{-1mm}
Domain Adaptation aims to transfer knowledge learned by a model from a source domain to a target domain, addressing issues such as data scarcity or high annotation costs in the target domain~\cite{farahani2021brief,csurka2017domain}. Traditional methods typically rely on adversarial training~\cite{tzeng2017adversarial,long2018conditional} or self-supervised learning~\cite{sun2019unsupervised,saito2020universal}. 
% In recent years, generative models, such as GANs~\cite{goodfellow2014generative} and diffusion models~\cite{ho2020denoising}, have shown potential in domain adaptation due to their ability to generate synthetic data or features for the target domain. 
For instance, some methods use the maximum mean discrepancy loss~\cite{wang2021rethinking,dziugaite2015training} to measure the divergence between different domains. In addition, the central moment discrepancy loss~\cite{zellinger2017central,xiong2021multi} and maximum density divergence loss~\cite{li2020maximum} are also proposed to align the feature distributions. Secondly, the domain adversarial training methods~\cite{ganin2016domain} learn the domain-invariant representations to encourage samples from different domains to be non-discriminative with respect to the domain labels via an adversarial loss. The third type of approach aims to minimize the cost transferred from the source to the target domain by finding an optimal cost to mitigate the domain shift~\cite{courty2017joint}. 

\subsection{Diffusion Models in Dense Prediction Tasks}
\vspace{-1mm}
Diffusion models have made a significant impact in the field of image generation, producing images of exceptionally high quality~\cite{ho2022cascaded,zhu2023conditional}. Moreover, in dense prediction tasks, such as depth estimation, diffusion models have also secured a place by accurately modeling source domain data~\cite{zhang2024three,ji2024dpbridge}. Marigold~\cite{ke2024repurposing} and Lotus~\cite{he2024lotus} have achieved remarkable results in monocular depth estimation task by fine-tuning pre-trained Stable Diffusion models~\cite{rombach2022high}. \citet{ji2023ddp} and SegDiff~\cite{amit2021segdiff} have demonstrated excellent performance in semantic segmentation task. DiffBIR~\cite{lin2024diffbir} and FlowIE~\cite{zhu2024flowie} leverage the powerful generative diffusion prior to super-resolution. However, most of the research focuses on the performance of their seen domain. The challenge of adapting diffusion-based dense prediction model to different domains remains an open and unresolved issue.
\vspace{-2mm}

\begin{figure}[htbp]
    \centering

    \begin{subfigure}[b]{0.45\textwidth}
        \includegraphics[width=\textwidth]{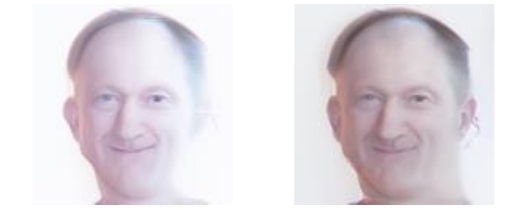}
        \caption{Visualization of Exposure Bias in ADM~\cite{bao2022analytic} on image generation} 
        \label{fig:visual_eb}
    \end{subfigure}
    
    \begin{subfigure}[b]{0.45\textwidth}
        \includegraphics[width=\textwidth]{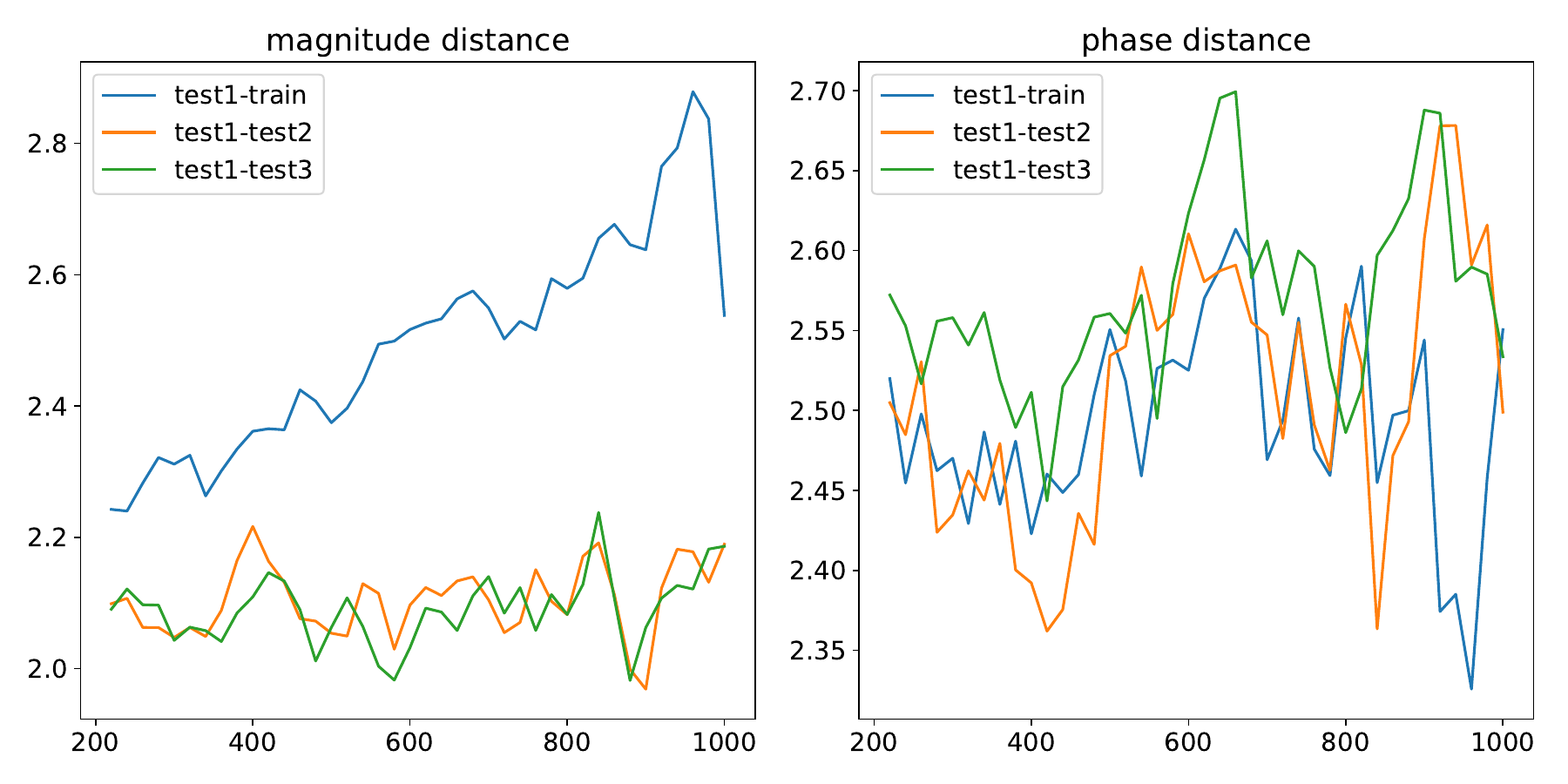}
        \caption{Exposure bias in ADM~\cite{bao2022analytic} on image generation}
        \label{fig:eb_adm}
    \end{subfigure}

    \begin{subfigure}[b]{0.45\textwidth}
        \includegraphics[width=\textwidth]{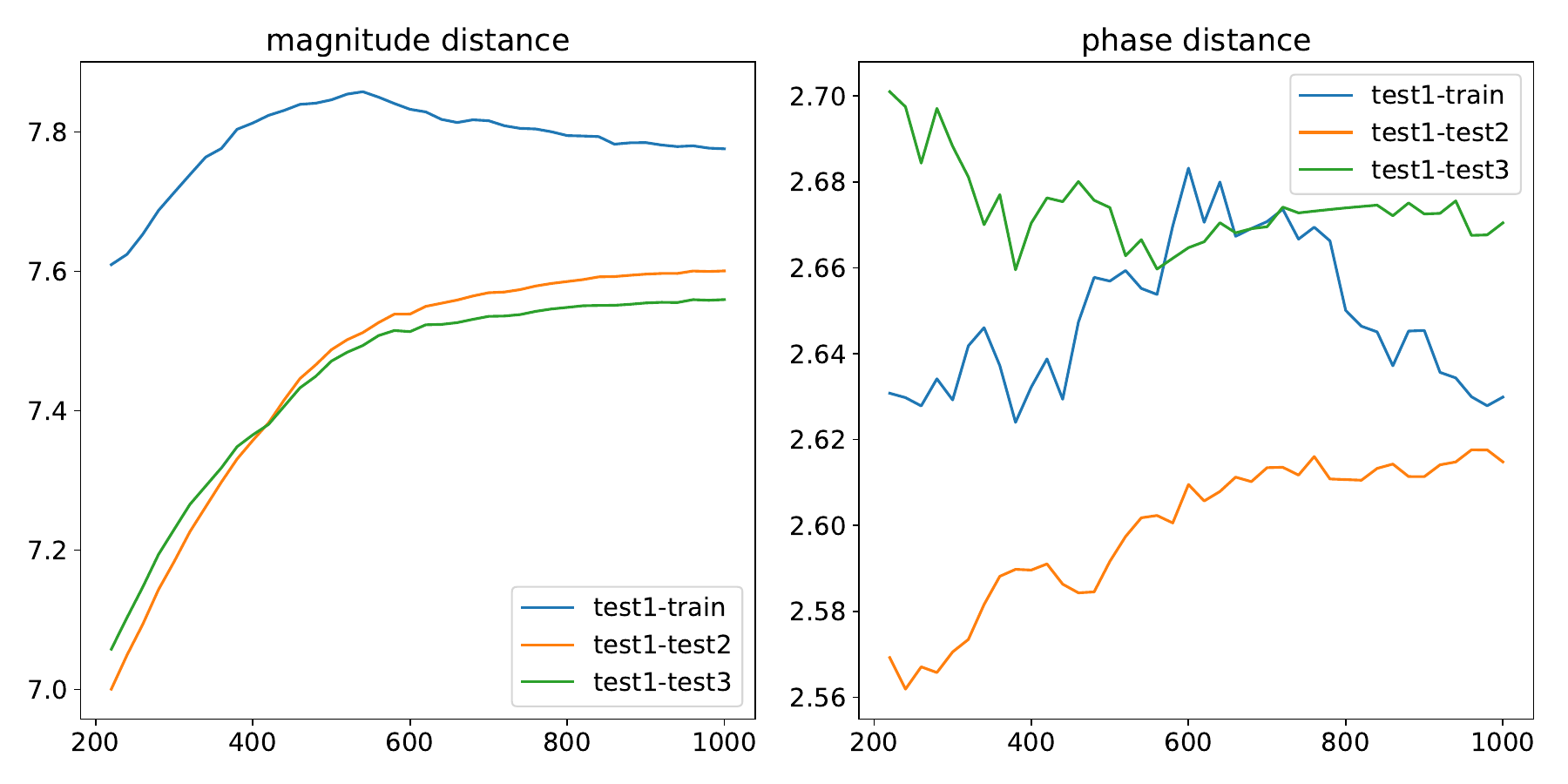}
        \caption{Exposure bias in Marigold~\cite{ke2024repurposing} on depth estimation}
        \label{fig:eb_marigold}
    \end{subfigure}
    
    \caption{(a) shows the visual results before(left) and after(right) addressing exposure bias. (b) and (c) visualize the amplitude differences of noise prediction between training and testing both in diffusion-based generation and DDP, further illustrating the relationship between exposure bias and domain shift. The legend "test 1" "test 2" and "test 3" represents the subsets we use to simulate the sampling process, and "train" represents the subset corresponding to "test 1" that is used to simulate the training process. Since amplitude is more related to domain information, the train-test gap brought by exposure bias is highly related to domain bias.}
    \vspace{-5mm}
    \label{fig:eb_distance}
\end{figure}
\begin{figure}[t]
	\centering
	\includegraphics[width=1\linewidth]{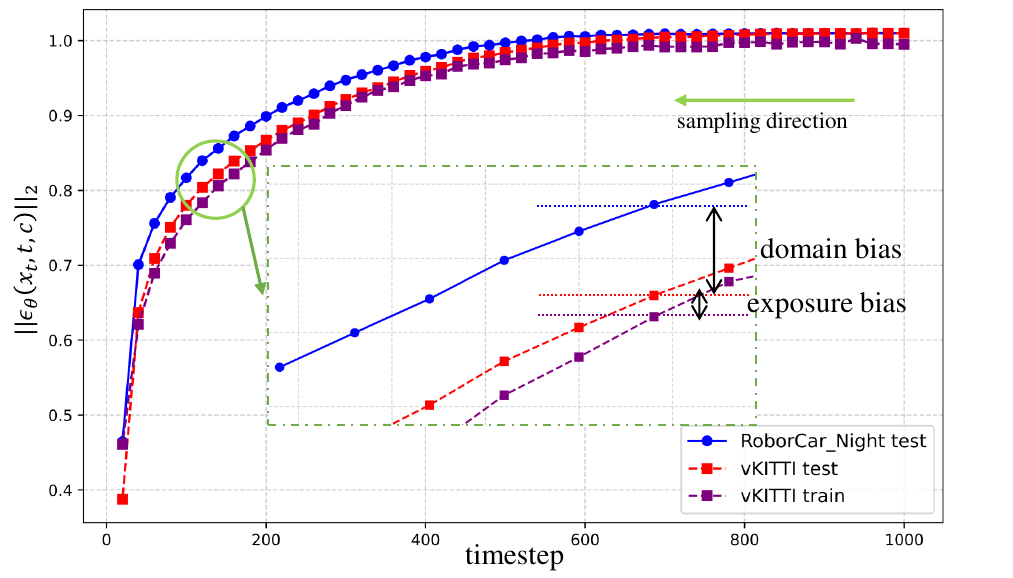}
	\caption{We observe that the prediction error brought by exposure bias primarily manifests as phase discrepancies, which suggests that exposure bias can be interpreted as domain shift. From this perspective, we discovered that in i2i diffusion models, the noise predictions between different domains exhibit phenomena similar to exposure bias, which we refer to as domain bias.  } 
	\label{fig:bias}
	\vspace{-10pt}
\end{figure}
\subsection{Diffusion Models for Domain Adaptation}
\vspace{-1mm}
Diffusion models excel at modeling domain attributes, which has led many researchers to explore diffusion-based domain adaptation~\cite{benigmim2023one,song2022diffusion}. However, diffusion models are primarily utilized as generators of target domain images~\cite{benigmim2023one} or tools for expanding the source domain~\cite{niemeijer2024generalization,niemeijer2024generalization}. For instance, DATUM~\cite{benigmim2023one} and DIDEX~\cite{niemeijer2024generalization} both leverage pre-trained diffusion models to augment the source domain, followed by other DA methods to enhance the robustness of the vision task models. In our work, our goal is to improve the performance of DDP models in the target domain, an area that remains largely unexplored in current research.

\section{Motivation}
\label{sec:motivation}
\subsection{Preliminary: Exposure Bias (EB)}
Exposure bias in diffusion models refers to the discrepancy that arises during sampling predictions due to the inability to access the ground truth noisy samples $\mathbf{x}_{t}$~\cite{ren2024multi,ning2023input}. Specifically, during training, we can access the ground truth $\mathbf{x}_{t}$ with forward process $q(x_t|x_0)$. During sampling, due to lack of $x_0$, we can only estimate $x_t$ with $q_\theta(\hat{x}_t|x_{t+1})$, leading to the prediction error in a single step. More severely, this error can accumulate along the sampling trajectory~\cite{ning2023elucidating}.  
% \citet{ning2023input} introduces a training regularisation term to simulate the sampling prediction errors from the Lipschitz continuity perspective. 
Recently, \citet{ning2023elucidating} proposes a training-free method called Noise Scaling that scales the epsilon predicted to fit the statistics during training. The scaling operation is:
\begin{equation}
\vspace{-3pt}
\label{eq:noise_scaling}
    \mu_{\boldsymbol{\theta}}(\boldsymbol{x}_t, t) = \frac{1}{\sqrt{\alpha_t}} \left( \boldsymbol{x}_t - \frac{\beta_t}{\sqrt{1 - \alpha_t}} \frac{\boldsymbol{\epsilon}_{\boldsymbol{\theta}}(\boldsymbol{x}_t, t,c)}{\lambda_t} \right).
\vspace{-3pt}
\end{equation}
Where $\lambda_t$ represents the scaling coefficient.

\subsection{From Exposure Bias to Domain Bias}
Noise Scaling~\cite{ning2023elucidating} alleviates the exposure bias by aligning the statistics of noise prediction between training and sampling, similar to AdaIN~\cite{huang2017arbitrary} in Style Transfer tasks. Moreover, ZeroSNR~\cite{lin2024common} addresses the inconsistency between training and testing by adjusting the noise schedule, helping diffusion models generate images with extreme brightness, which is highly relevant to domain style. All of these points implicitly illustrate the relationship between exposure bias and domain. We then further investigate the differences in noise prediction between training and testing from a Fourier perspective, as shown in~Fig.~\ref{fig:eb_distance}(b)(c). In the amplitude spectrum, there is a significant difference between the noise predictions during training and testing, while in the phase spectrum, such differences are difficult to observe. Moreover, exposure bias leads to visual changes in~Fig.~\ref{fig:eb_distance}(a).  These phenomena suggest that exposure bias can be understood as domain shift. With this insight, we aim to explore whether a phenomenon similar to exposure bias also exists in diffusion models when encountering a true domain gap. 

Let's consider a conditional diffusion model $\boldsymbol{\epsilon}_{\boldsymbol{\theta}}(\boldsymbol{x}_t, t,c)$ that has been pre-trained on image translation tasks. Given a frozen model and timestep $t$, when condition $c$ comes from the source domain $D^s$, we can collect the output noise $\boldsymbol{\epsilon}_{\boldsymbol{\theta}}^s$ by $\boldsymbol{\epsilon}_{\boldsymbol{\theta}}(\boldsymbol{x}_t, t,c_s)$. Similarly, we can collect the output noise $\boldsymbol{\epsilon}_{\boldsymbol{\theta}}^t$ by $\boldsymbol{\epsilon}_{\boldsymbol{\theta}}(\boldsymbol{x}_t, t,c_t)$ when condition c comes from the target domain $D^t$. One can infer that the prediction $\boldsymbol{\epsilon}_{\boldsymbol{\theta}}^s$ is always more reliable than the prediction $\boldsymbol{\epsilon}_{\boldsymbol{\theta}}^t$, given that the model has never been trained on the target domain, and there exists a significant bias between them, as proposed in Fig.~\ref{fig:bias}. We refer to this bias as domain bias, which represents the differences in noise prediction $\boldsymbol{\epsilon}_{\boldsymbol{\theta}}(\boldsymbol{x}_t, t,c)$ with conditions from different domains. When the domain gap between the target domain and the source domain is substantial, the domain bias significantly outweighs the exposure bias, allowing us to focus solely on addressing the domain bias.

Domain bias and exposure bias share several commonalities: their existence stems from the gap between the unreliable predictions and the reliable predictions of the network. Therefore, we can refer to the operation in addressing exposure bias (i.e., adjusting the noise prediction statistics) as a potential solution to address the domain bias in DDP.

\begin{figure}[t]
	\centering
	\includegraphics[width=1\linewidth]{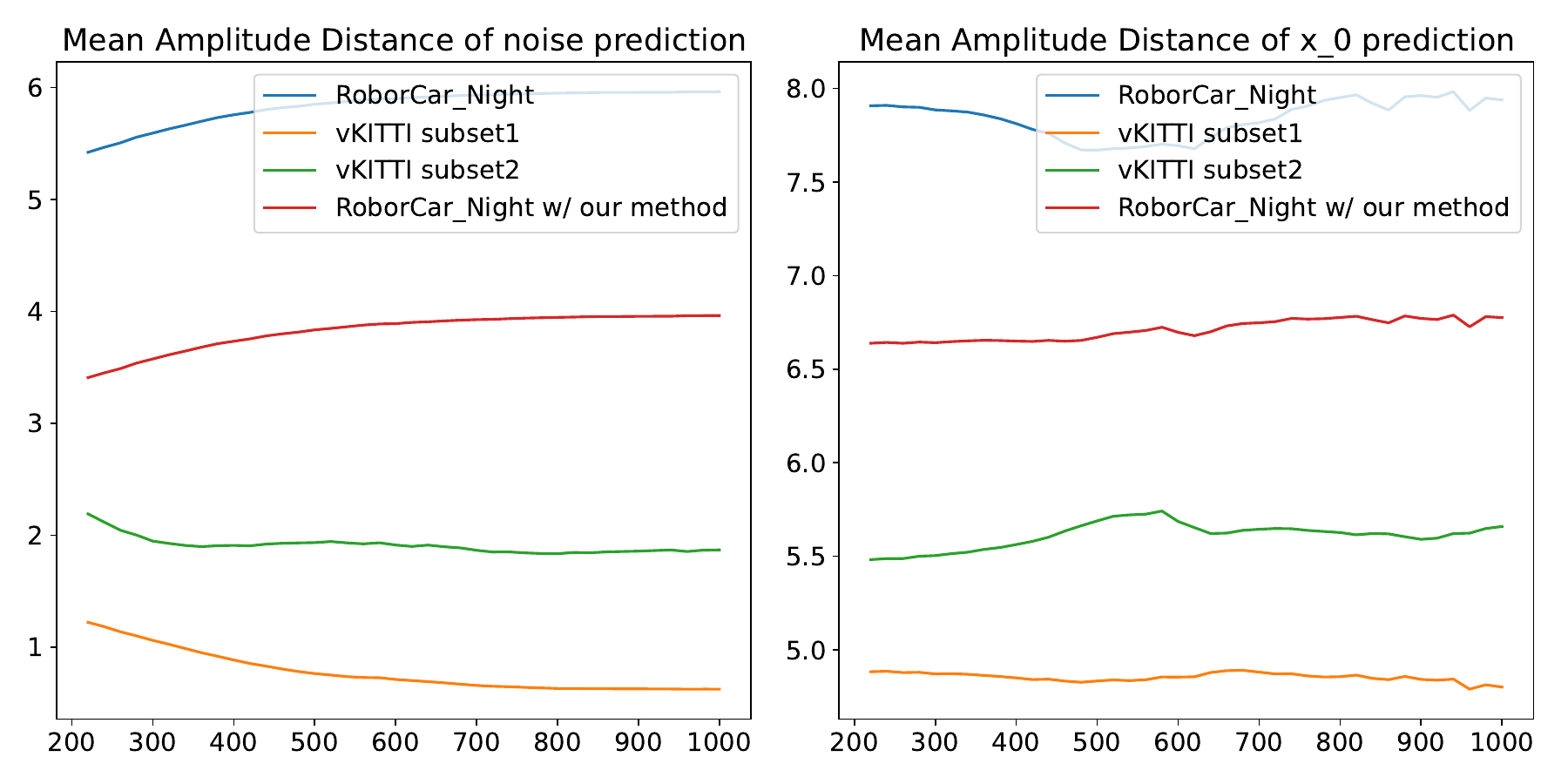}
 	\vspace{-5pt}
	\caption{ \textbf{The amplitude difference between different domains in timestep $t$.} The amplitude difference between noise prediction and $x_0$ prediction remains highly consistent, indicating that adjusting the statistics of noise can help reduce style bias and domain bias, given the strong correlation between amplitude and style. }
	\label{fig:fourier}
	\vspace{-10pt}
\end{figure}

\section{Method}
\label{sec:method}
\subsection{EB-inpired scaling for source-available DA}
\label{subsec:sada}
To alleviate the domain bias, we propose Domain Noise Alignment, where we adjust the sampling variance of $\boldsymbol{\epsilon}_{\boldsymbol{\theta}}^s$ to shift the model's predictions from the unreliable vector field of the target domain to the reliable vector field of the source domain. To analyze why Domain Noise Alignment works, we first explain that the high stylistic consistency between noise prediction $\boldsymbol{\epsilon}_{\boldsymbol{\theta}}$, estimated output $\hat{x}_0$, and the condition $c$. Taking the depth estimation task as an example, we select a subset $A$ from the training set vKITTI~\cite{gaidon2016virtual} as the baseline. Two additional subsets, $B$ and $C$, are chosen for reference. We then select the test set RobotCar-Night~\cite{maddern20171} as the target domain. Therefore, domains $B$ and $C$ naturally exhibit smaller differences compared to domain $A$, while the test set shows more significant variations. We only need to focus on $\boldsymbol{\epsilon}_{\boldsymbol{\theta}}$ and $\hat{x}_0$.  We use the amplitude difference after Fourier transformation to represent stylistic differences. After applying Fourier transformation to the noise prediction at each step and the depth map predicted at each step, we compare their amplitude with the baseline subset $A$, as shown in~Fig.~\ref{fig:fourier}, which strongly validates our explanation. Therefore, the stylistic information of the noise prediction $\boldsymbol{\epsilon}_{\boldsymbol{\theta}}$ also represents the stylistic information of the condition $c$ and the output images $x_0$. Moreover, the first and second-order statistical moments of a distribution are crucial indicators of its stylistic characteristics. So, adjusting the variance of the noise can effectively align with the stylistic features. Specifically, assuming we have an accurate domain converter $G$ that satisfies $G(D^t)=D^s$, then $\boldsymbol{\epsilon}_{\boldsymbol{\theta}}^{G(t)}$=$\boldsymbol{\epsilon}_{\boldsymbol{\theta}}(\boldsymbol{x}_t, t, G(c_t))$ should be a reliable prediction. Unfortunately, we do not have such a converter, so we need to estimate $\boldsymbol{\epsilon}_{\boldsymbol{\theta}}^{G(t)}$ with $\boldsymbol{\epsilon}_{\boldsymbol{\theta}}^{t}$. Considering that $G(c_t)$ and $c_s$ belong to the same domain, and the noise prediction is stylistically consistent with the condition, we can reasonably assume that $\boldsymbol{\epsilon}_{\boldsymbol{\theta}}^{G(t)}$ and $\boldsymbol{\epsilon}_{\boldsymbol{\theta}}^{s}$ are stylistically similar. Meanwhile, style information can be represented using first- and second-order statistics, so we can assume that $var(\boldsymbol{\epsilon}_{\boldsymbol{\theta}}^{G(t)}) \approx var(\boldsymbol{\epsilon}_{\boldsymbol{\theta}}^{s})$. Therefore, although $G(c_t)$ is not accessible, We can align the statistics of $\boldsymbol{\epsilon}_{\boldsymbol{\theta}}^{t}$ with those of $\boldsymbol{\epsilon}_{\boldsymbol{\theta}}^{s}$, thereby approximately matching with the style information of $\boldsymbol{\epsilon}_{\boldsymbol{\theta}}^{G(t)}$. Furthermore, considering that $\boldsymbol{\epsilon}_{\boldsymbol{\theta}}^{t}$ and $\boldsymbol{\epsilon}_{\boldsymbol{\theta}}^{G(t)}$ are inherently semantically consistent, the adjusted $\boldsymbol{\epsilon}_{\boldsymbol{\theta}}^{t}$ serves as a good estimate for $\boldsymbol{\epsilon}_{\boldsymbol{\theta}}^{G(t)}$.

\begin{figure}[t]
    \centering
    \begin{subfigure}{0.23\textwidth}
        \includegraphics[width=\linewidth]{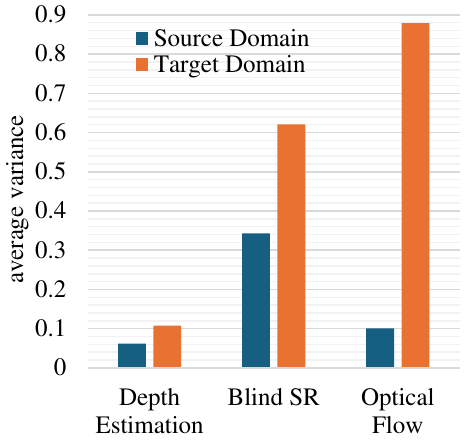}
        \caption{Average variance of multiple outputs in different tasks.}
        \label{fig:left}
    \end{subfigure}
    \hfill
    \begin{subfigure}{0.22\textwidth}
        \includegraphics[width=\linewidth]{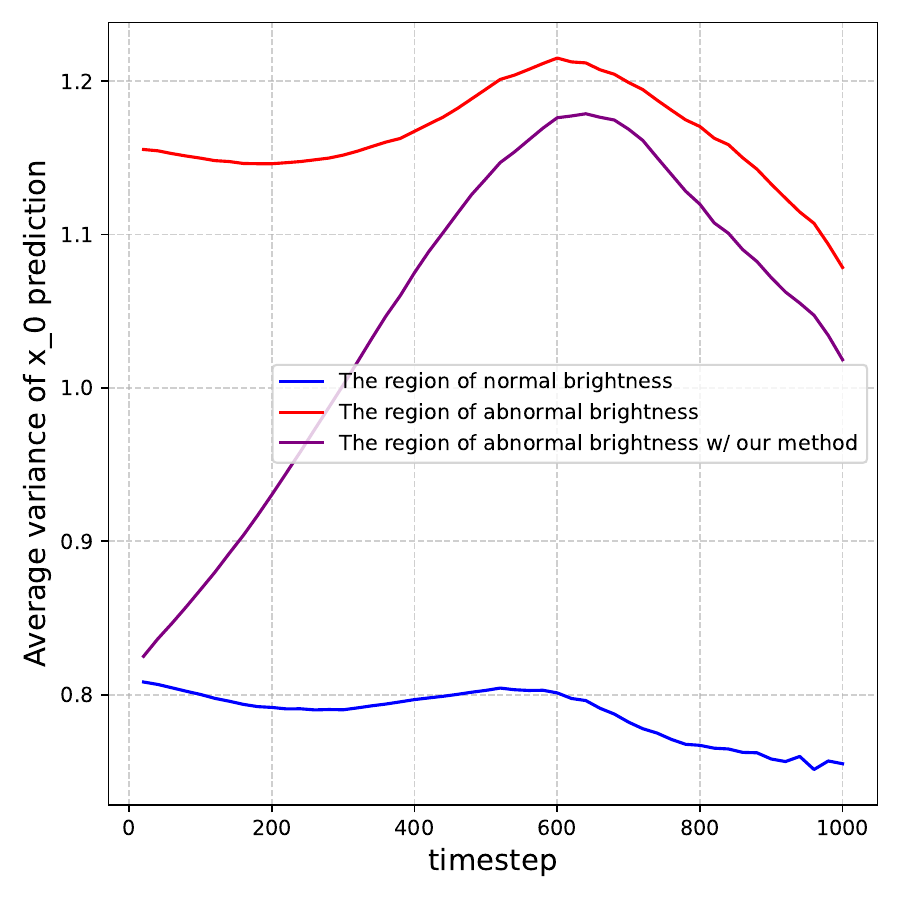}
        \caption{Average variance of  outputs in regions of varying lighting.}
        \label{fig:right}
    \end{subfigure}
    \caption{(a) illustrates that the DDP is more likely to produce consistent predictions (lower variance) when the conditional image is close to the source domain, where the consistency is measured by averaging the variance across multiple outputs at each pixel location. (b) taking Marigold trained on the normal-light dataset and test on uneven abnormal-light dataset as example, where the model struggles to produce consistent results in areas with abnormal lighting, which can be effectively alleviated by our method.}
    \label{fig:consistency}
    \vspace{-10pt}
\end{figure}

Directly aligning the variance of $\boldsymbol{\epsilon}_{\boldsymbol{\theta}}^{t}$ and $\boldsymbol{\epsilon}_{\boldsymbol{\theta}}^{s}$ can lead to excessive adjustments due to the ongoing impact on subsequent steps. A more reasonable approach is to utilize the scaling schedule proposed by~\citet{ning2023elucidating} to calculate the $\lambda_t$ in Eq.\ref{eq:noise_scaling} for each timestep: $\Delta N(t) - 1 \approx \int_{t}^{T} (\lambda_T - 1) dt + \int_{t}^{T-1} (\lambda_{T-1} - 1) dt + ... + \int_{t}^{t+1} (\lambda_{t+1} - 1) dt$, where $\Delta N(t) = \|\boldsymbol{\epsilon}_{\boldsymbol{\theta}}(\boldsymbol{x}_t, t,c_s)\|_2/\|\boldsymbol{\epsilon}_{\boldsymbol{\theta}}(\boldsymbol{x}_t, t,c_t)\|_2$. We can approximate $\lambda_t$ by leveraging the error between two adjacent timesteps:
\begin{equation}
\vspace{-3pt}
\label{eq:chafen}
    \Delta N(t) - \Delta N(t+1) = \sum\nolimits_{t+2}^{T}{(\lambda_t-1)}+\lambda_{t+1}-1.
\vspace{-5pt}
\end{equation}

\begin{equation}
\vspace{-3pt}
\label{eq:cal_lambda_t}
   \lambda_t \approx \lambda_{t+1} = \Delta N(t) - \Delta N(t+1)+1-\sum\nolimits_{t+2}^{T}{(\lambda_t-1)}.
\vspace{-3pt}
\end{equation}

\begin{figure}[t]
	\centering
 
    \begin{subfigure}[b]{0.45\textwidth}
        \includegraphics[width=1\linewidth]{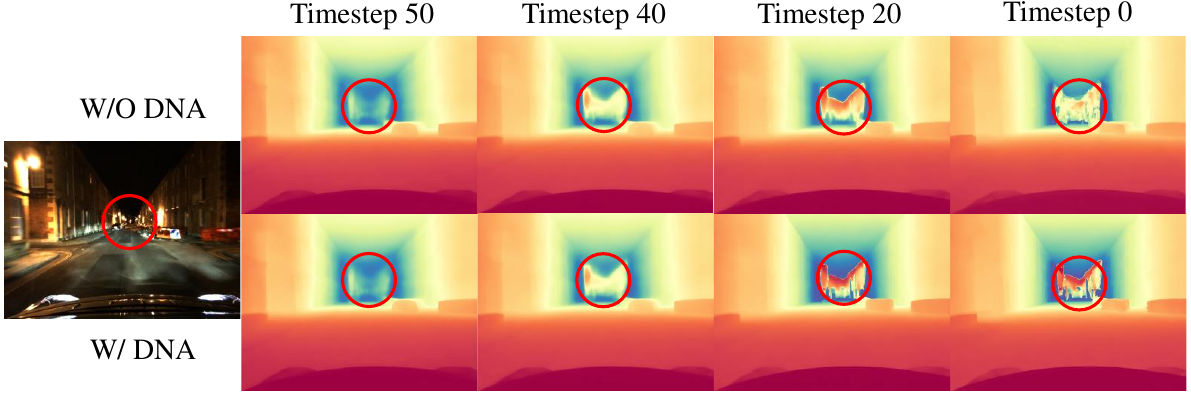}
	  \caption{The progressive adjustment for the low-confidence region. With our method, diffusion models adjust their predictions for these regions toward a more accurate direction(\textcolor{red}{red circles}).} 
	  \label{fig:inter_step}
    \end{subfigure}
    
    \begin{subfigure}[b]{0.45\textwidth}
        \includegraphics[width=\textwidth]{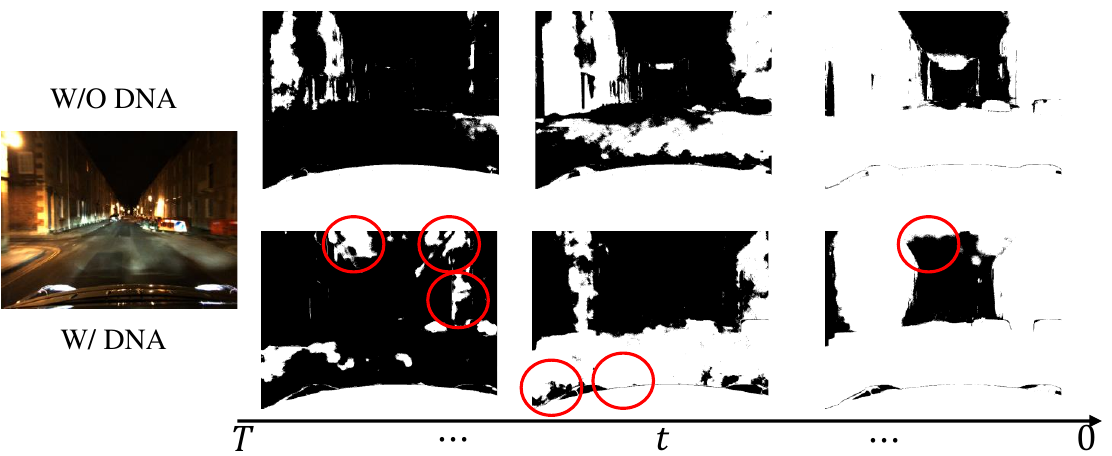}
        \caption{The variation of high-confidence region mask along the timestep in Marigold trained on the normal-light dataset. In the first row, the mask covers almost all areas with the most normal lighting. With our DNA, many areas with extremely poor lighting also achieve more consistent prediction results, some of which are highlighted with \textcolor{red}{red circles}.} 
        \label{fig:mask}
    \end{subfigure}
    \vspace{-5pt}
    \caption{Visualization of DNA with progressive adjustment.}
    \label{fig:progressive}
	\vspace{-10pt}
\end{figure}

\subsection{Noise scaling estimation for source-free DA}
\label{subsec:sfda}
The calculation of $\lambda_t$ depends on $\boldsymbol{\epsilon}_{\boldsymbol{\theta}}(\boldsymbol{x}_t, t,c_s)$, meaning that the distribution of the source domain $D^s$ must be accessible. However, when the source domain $D^s$ is inaccessible, $\lambda_t$ becomes difficult to compute. Fortunately, we can estimate the source domain's distribution with a good medium — multiple noise sampling. We have observed that the closer the conditional image is to the source domain, the more consistent the results are likely to be when the initial noise varies, as shown in~Fig.~\ref{fig:consistency}(a). Therefore, by performing multiple sampling of the initial noise in a batch, we can generate target images $x_0 \in R^{B\times C\times H \times W}$ and define high-confidence and low-confidence regions based on their variances along the batch dimension, where $B$ represents the number of initial noise samples. The regions with lower variance are more reliable compared to other regions. Therefore, we use $p$ as the percentage threshold to obtain regions with higher confidence and approximate the statistical properties of the source domain during the denoising process by leveraging the statistics of these selected regions:
\begin{equation}
\vspace{-3pt}
\label{eq:threshold}
    M_h = |var(x_0)|<quantile(var(x_0),p),
\vspace{-2pt}
\end{equation}
\begin{equation}
\vspace{-1pt}
\label{eq:delta}
    \Delta N(t) = \|\boldsymbol{\epsilon}_{\boldsymbol{\theta}}(\boldsymbol{x}_t, t,c_s)\|_2/\|\boldsymbol{\epsilon}_{\boldsymbol{\theta}}(\boldsymbol{x}_t, t,c_s)[M_h]\|_2.
\vspace{-2pt}
\end{equation}

\begin{figure}[t]
	\centering
	\includegraphics[width=1\linewidth]{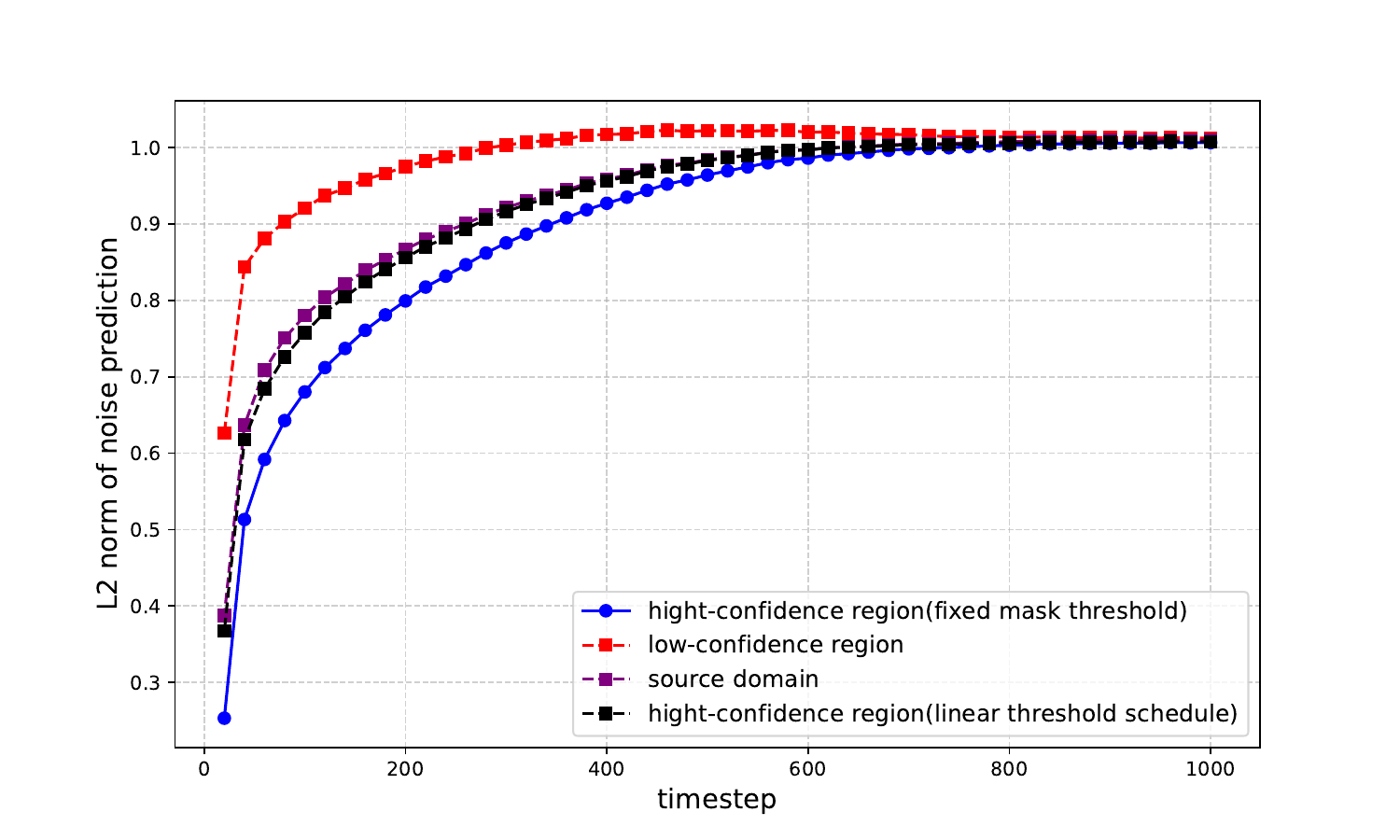}
	\caption{Based on the results obtained from multiple samplings, we divide the regions into high-confidence and low-confidence areas. It is observed that the variance of noise in high-confidence regions(\textcolor{blue}{blue line}) is always closer to that of the source domain's noise prediction(\textcolor{purple}{purple line}), compared with that in low-confidence regions(\textcolor{red}{red line}). Considering that the accuracy of high-confidence regions varies at different timesteps, we employ a linear threshold schedule and refine $\lambda_t$ by excluding the factor of noise consistency in certain regions, achieving a more accurate estimation of the source domain's noise statistics(\textcolor{black} {black line}).} 
	\label{fig:confidence}
	\vspace{-10pt}
\end{figure}

Then DNA is adopted to adjust the noise prediction progressively, the effectiveness of which is shown in Fig.~\ref{fig:progressive}(a). The model’s predictions for low-confidence regions increasingly align with the output of the source domain, as illustrated in Fig.~\ref{fig:consistency}(b), thereby enabling more accurate predictions. Meanwhile, the gap between low-confidence and high-confidence regions is gradually narrowing, as shown in Fig.~\ref{fig:progressive}(b).
To enhance the accuracy and efficiency of estimating the statistical properties of the source domain, we have introduced three additional designs to improve efficiency and achieve accurate estimation, as shown in~Fig.~\ref{fig:confidence}.

\noindent\textbf{Online Noise Alignment.} At each timestep, we can derive an approximate estimation of $x_0$ and through $x_0 = (\boldsymbol{x}_t-\sqrt{\bar{\beta}_t}\boldsymbol{\epsilon}_{\boldsymbol{\theta}}(\boldsymbol{x}_t, t,c_s))/\sqrt{\bar{\alpha}_t}$. Although this estimation may not be entirely accurate, it still allows us to obtain a rough high-confidence region, which serves as an approximation of the source domain and helps avoid the pre-calculation of $M_h$.

\noindent\textbf{Noise Consistency Scaling.} Region of high confidence may result from the high consistency of noise in some regions. For convenience, we simply divide the noise into four regions and calculate the consistency of the noise within each batch in different regions respectively. We perform calculations on the noise predicted at each timestep and use the pdist function (pairwise distance in n-dimensional space) as an indicator of consistency. Although the impact is minimal, removing the factor of $\gamma$ from $\lambda_t$ can lead to a slight improvement.
\begin{table}[htbp]
\centering
\caption{Ablation on mask schedule for depth estimation.}
\label{tab:mask}
\begin{tabular}{lcc}
\toprule
\multirow{2}{*}{Method} & \multicolumn{2}{c}{Marigold~RobotCar\_Night} \\
\cmidrule(lr){2-3}
& AbsRel $\downarrow$ & $\delta_1$ $\uparrow$ \\ \hline
same in all steps &19.4 &68.1 \\
larger in earlier steps &21.1 &65.7 \\
\textbf{larger in later steps} &\textbf{18.2} &\textbf{71.4} \\
\bottomrule
\end{tabular}
\end{table}

\noindent\textbf{Linear Threshold Schedule.} We obtain the mask $M_h$ with the percentage threshold $p$ during sampling. However, the high-confidence thresholds corresponding to different steps should vary. Many previous UDA models have adopted a progressive optimization strategy for pseudo-labels~\cite{chen2019progressive,kim2021semi,li2022unsupervised}. Inspired by this, we believe that high-confidence regions should be smaller in the early denoising steps and larger in the later denoising steps, which is also demonstrated in~Tab.~\ref{tab:mask}. Therefore, we model $p$ as a linear function of $t$ to avoid the complex search for thresholds, which shows effectiveness in~Fig.~\ref{fig:confidence}.

% \begin{table}[htbp]
% \centering
% \caption{Comparison results of diffusion models w/ and w/o our methods for depth estimation under Source-Available DA setting.}
% \label{tab:nyuv2}
% \begin{tabular}{lccccc}
% \toprule
% \multirow{2}{*}{Method} & \multicolumn{2}{c}{NuScenes} & \multicolumn{2}{c}{RoborCar\_Night} \\
% \cmidrule(lr){2-3} \cmidrule(lr){4-5}
%  & AbsRel $\downarrow$ & $\delta_1$ $\uparrow$ & AbsRel $\downarrow$ & $\delta_1$ $\uparrow$ \\
% \midrule
% Marigold &  &  & 23.8 & 63.2 \\
% Marigold+ours &  &  & 17.0 & 74.9 \\ \hline
% Lotus &  &  & 22.1 & 65.9 \\
% Lotus+ours &  &  & 17.2 & 78.1 \\
% \bottomrule
% \end{tabular}
% \end{table}

% \begin{table}[htbp]
% \centering
% \caption{Comparison results of diffusion models w/ and w/o our methods for depth estimation under source-free DA setting.}
% \label{tab:nyuv2}
% \begin{tabular}{lccccc}
% \toprule
% \multirow{2}{*}{Method} & \multicolumn{2}{c}{NuScenes} & \multicolumn{2}{c}{RoborCar\_Night} \\
% \cmidrule(lr){2-3} \cmidrule(lr){4-5}
%  & AbsRel $\downarrow$ & $\delta_1$ $\uparrow$ & AbsRel $\downarrow$ & $\delta_1$ $\uparrow$ \\
% \midrule
% Marigold &  &  & 23.8 & 63.2 \\
% Marigold+ours &  &  & 18.2 & 71.4 \\ \hline
% Lotus &  &  & 22.1 & 65.9 \\
% Lotus+ours &  &  & 18.6 & 71.3 \\
% \bottomrule
% \end{tabular}
% \end{table}

\begin{table}[htbp]
\centering
\caption{Comparison results of diffusion models w/ and w/o our methods for depth estimation under different DA settings.}
\label{tab:exp_depth}
\resizebox{0.45\textwidth}{!}{
\begin{tabular}{lccccc}
\toprule
\multirow{2}{*}{Method} & \multicolumn{2}{c}{NuScenes} & \multicolumn{2}{c}{RoborCar\_Night} \\
\cmidrule(lr){2-3} \cmidrule(lr){4-5}
 & AbsRel $\downarrow$ & $\delta_1$ $\uparrow$ & AbsRel $\downarrow$ & $\delta_1$ $\uparrow$ \\ \hline
\multicolumn{5}{c}{Source-Available Domain Adaptation} \\ \hline
S2R-DepthNet~\cite{chen2021s2r} &47.4  &40.8 &32.1 &55.2 \\
DESC~\cite{lopez2023desc} &45.8  &39.4   &30.6 &55.1 \\ \hline
Marigold~\cite{ke2024repurposing} &33.7  &50.1  & 23.8 & 63.2 \\
\textbf{Marigold+ours} &\textbf{26.3} &\textbf{61.5} &\textbf{17.0} &\textbf{74.9} \\ \hline
Lotus~\cite{he2024lotus} &30.2  &54.8  & 22.1 & 65.9 \\
\textbf{Lotus+ours} &\textbf{26.4} &\textbf{62.2} &\textbf{17.2} &\textbf{78.1} \\ \hline
\multicolumn{5}{c}{Source-Free Domain Adaptation} \\ \hline
Ada-depth~\cite{li2023test} &48.5 &37.7 & 34.9 &51.4 \\ \hline
Marigold &33.7  &50.1  & 23.8 & 63.2 \\
\textbf{Marigold+ours} &\textbf{28.8} &\textbf{57.0} &\textbf{18.2} &\textbf{71.4} \\ \hline
Lotus &30.2  &54.8  & 22.1 & 65.9 \\
\textbf{Lotus+ours} &\textbf{27.5} &\textbf{60.6} &\textbf{18.6} &\textbf{71.3} \\
\bottomrule
\end{tabular}}
\end{table}
\vspace{-5pt}

\section{Experiments}
\label{sec:experiments}
\subsection{Depth Estimation}
\noindent\textbf{Implementation.} We adopt Marigold~\cite{ke2024repurposing} and Lotus~\cite{he2024lotus} as the baseline and set the timestep to 50 for inference. We employed two widely used depth estimation metrics to evaluate performance~\cite{ranftl2021vision,ranftl2020towards}. The first is the Absolute Mean Relative Error (AbsRel), calculated as: $\frac{1}{M} \sum_{i=1}^{M} \frac{|a_i - d_i|}{d_i}$, where $M$ is the total number of pixels. The second metric, $\delta_1$ accuracy, which is calculated by $\max\left(\frac{a_i}{d_i}, \frac{d_i}{a_i}\right) < 1.25 $.

\noindent\textbf{Dataset.} For Marigold, it adopts Hypersim~\cite{roberts2021hypersim} and vKITTI~\cite{gaidon2016virtual} as training datasets. To widen the domain gap between the source and target domains, we utilized two nighttime depth estimation datasets: the NuScenes-night~\cite{caesar2020nuscenes} and RobotCar-night datasets~\cite{maddern20171}. We also conducted tests on other depth estimation datasets, such as NYUv2~\cite{silberman2012indoor}, the results of which are shown in Supplementary materials. 

\noindent\textbf{Performance.} We present the results of Marigold and Lotus with our methods compared with previous UDA models~\cite{chen2021s2r,lopez2023desc,li2023test} under different DA settings in Tab.~\ref{tab:exp_depth}. Our proposed method has surpassed all previous methods across both metrics, even though our approach is training-free. In Fig.~\ref{fig:confidence}, we present the visual comparison, revealing that our method achieves more accurate estimations in areas with lower illumination (further from the source domain).

\begin{table}[htbp]
\centering
\caption{Comparison results of diffusion models w/ and w/o our methods for Blind SR under different DA settings.}
\label{tab:exp_bsr}
\resizebox{0.45\textwidth}{!}{
\begin{tabular}{lccccc}
\toprule
\multirow{2}{*}{Method} & \multicolumn{4}{c}{DRealSR}\\
\cmidrule(lr){2-5}
 & PSNR$\uparrow$ & LPIPS$\downarrow$ &MUSIQ$\uparrow$ &MANIQA$\uparrow$\\ \hline
\multicolumn{5}{c}{Source-Available Domain Adaptation} \\ \hline
DiffBIR~\cite{lin2024diffbir} &24.05  &0.498  &65.12 &0.55 \\
DiffBIR+ours &\textbf{25.19} &\textbf{0.485} &\textbf{65.44} &\textbf{0.57}\\ \hline
StableSR~\cite{wang2024exploiting} &24.23  &0.503  &62.47 & 0.51\\
StableSR+ours&\textbf{25.22} &\textbf{0.487} &\textbf{63.71} &\textbf{0.54}\\ \hline
\multicolumn{5}{c}{Source-Free Domain Adaptation} \\ \hline
DiffBIR~\cite{lin2024diffbir} &24.05  &0.498  &65.12 &0.55\\
DiffBIR+ours &\textbf{24.70} &\textbf{0.491} &\textbf{65.52} &\textbf{0.59}\\ \hline
StableSR~\cite{wang2024exploiting} &24.23  &0.503  &62.47 & 0.51\\
StableSR+ours &\textbf{24.65} &\textbf{0.494} &\textbf{63.70} &\textbf{0.51}\\
\bottomrule
\end{tabular}}
\end{table}
\subsection{Blind Super-Resolution}
\noindent\textbf{Implementation.} We adopt DiffBIR~\cite{lin2024diffbir} and StableSR~\cite{wang2024exploiting} as the baseline. For DiffBIR, we do not fix the CFG scale~\cite{ho2022classifier} during comparisons. This decision stems from the consideration that adjusting the noise distribution can influence the diffusion model's ability to generate images that faithfully adhere to the natural distribution, potentially leading to a reduction in non-reference metrics. Therefore, our goal is to align the noise distribution while simultaneously adjusting the CFG scale to achieve a balance between fidelity and realism. For StableSR, we set the timestep to 50 for inference. The performance of various methods is evaluated using the classical metrics: PSNR, SSIM, and LPIPS, while we still provide two non-reference metrics, Maniqa~\cite{yang2022maniqa} and MUSIQ~\cite{ke2021musiq} for reference.

\noindent\textbf{Dataset.} DiffBIR adopts Laion-2b-en~\cite{schuhmann2022laion} for training and StableSR adopts DIV2K~\cite{agustsson2017ntire} and Flickr2K~\cite{timofte2017ntire} for training. For inference, to evaluate our method on real-world scenes, we adopt two paired real-world datasets RealSR~\cite{cai2019toward} and DRealSR~\cite{wei2020component}. We show the results of DRealSR are shown in~Tab.~\ref{tab:exp_bsr} and the left is shown in supplementary materials. 

\noindent\textbf{Performance.} We present the results of DiffBIR and StableSR with our methods under different DA settings in Tab.~\ref{tab:exp_bsr}. Our method elevates the PSNR by over 0.5 dB, while also improving non-reference metrics. Although the pre-processing network (SwinIR~\cite{liang2021swinir}) can generate coarse HR images, it is trained on synthetic degradation and struggles with real degradation. Therefore, our method indirectly narrows the gap between synthetic and real degradation, thereby enhancing performance in real-world scenarios.
\begin{figure*}[t]
	\centering
	\includegraphics[width=1\linewidth]{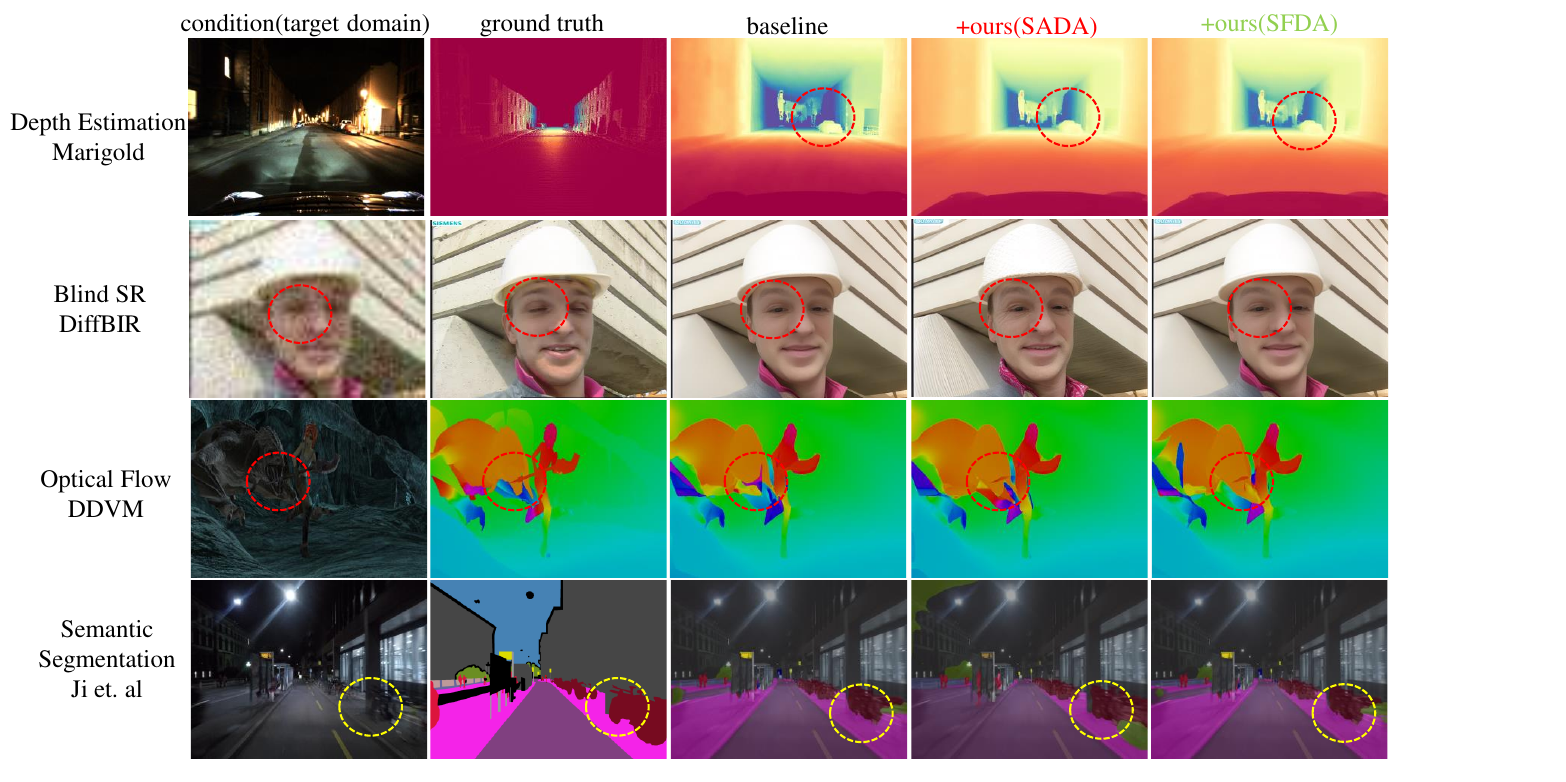}
	\caption{We have demonstrated the effectiveness of our approach across three tasks. Here, \textcolor{red}{+ours(SADA)} denotes the results obtained using our proposed Domain Noise Alignment under the Source-Available DA setting, \textcolor{green}{+ours(SFDA)} represents the results achieved with our proposed Source-Free Domain Noise Alignment under the Source-Free DA setting. The differences are highlighted with red circles.} 
	\label{fig:visual_results}
	\vspace{-10pt}
\end{figure*}

\begin{table}[htbp]
	\vspace{-5pt}
\centering
\caption{Comparison results of diffusion models w/ and w/o our methods for depth estimation under different DA settings.}
\label{tab:exp_flow}
\begin{tabular}{lccccc}
\toprule
\multirow{2}{*}{Method} & \multicolumn{2}{c}{FCDN} & \multicolumn{2}{c}{Sintel} \\
\cmidrule(lr){2-3} \cmidrule(lr){4-5}
 & EPE$\downarrow$ & 1px$\uparrow$ & EPE$\downarrow$ & 1px$\uparrow$ \\ \hline
\multicolumn{5}{c}{Source-Available Domain Adaptation} \\ \hline
TST~\cite{yoon2024optical} &8.18  &55.8  &3.87 &68.1  \\
UNDAF~\cite{wang2022undaf} &7.47  &55.3  & 4.02 & 72.3 \\ 
UCDA-Flow~\cite{zhou2023unsupervised} &6.34  &62.8  & 3.29 & 75.9 \\ \hline
DDVM~\cite{dong2023open} &7.41  &59.4  & 2.36 & 80.5 \\
DDVM+ours &\textbf{4.93} &\textbf{64.3} &\textbf{1.99} &\textbf{81.7} \\ \hline
\multicolumn{5}{c}{Source-Free Domain Adaptation} \\ \hline
TTA-MV~\cite{ayyoubzadeh2023test} &9.42  &49.1  & 4.52 & 63.4 \\ \hline
DDVM &7.41  &59.4  & 2.36 & 80.5 \\
DDVM+ours &\textbf{6.33} &\textbf{62.4} &\textbf{2.06} &\textbf{80.6} \\
\bottomrule
\end{tabular}
	\vspace{-5pt}
\end{table}
\subsection{Optical Flow}
\noindent\textbf{Omplementation} We adopt Open-DDVM~\cite{dong2023open} as the baseline. We choose the average end-point error (EPE) and the lowest percentage of erroneous pixels (Fl-all) as the quantitative evaluation metrics. For optical flow, it takes two RGB images as condition, and we set timestep to 50 for inference.

\noindent\textbf{Dataset.} Open-DDVM adopts AutoFlow~\cite{sun2021autoflow} for training. For inference, we would like to widen the domain gap between the source domain and the target domain. Therefore, we adopt the FCDN datasets~\cite{zheng2020optical}. We also provide the performance on the commonly used Sintel dataset~\cite{butler2012naturalistic}.

\noindent\textbf{Performance.} We present the results of Open-DDVM with our methods compared with previous UDA models under different DA settings in Tab.~\ref{tab:exp_bsr}. On both nighttime optical flow datasets, our methods surpass previous UDA models~\cite{yoon2024optical,wang2022undaf,zhou2023unsupervised,ayyoubzadeh2023test} across two metrics. Without training, Open-DDVM with our methods can generate more accurate optical flow given dark images as conditions.

\begin{table}[htbp]
\centering
\caption{Comparison results of diffusion models w/ and w/o our methods for semantic segmentation under different DA settings.}
\label{tab:exp_seg}
\resizebox{0.45\textwidth}{!}{
\begin{tabular}{lccccc}
\toprule
\multirow{2}{*}{Method} & \multicolumn{2}{c}{ACDC} & \multicolumn{2}{c}{DarkZurich} \\
\cmidrule(lr){2-3} \cmidrule(lr){4-5}
 & aAcc$\uparrow$ & mIOU$\uparrow$ & aAcc$\uparrow$ & mIOU$\uparrow$ \\ \hline
\multicolumn{5}{c}{Source-Available Domain Adaptation} \\ \hline
DACS~\cite{tranheden2021dacs} &0.76 &40.1 &- &- \\
DAFormer~\cite{hoyer2022daformer} &0.86 &55.4 &- &- \\ \hline
~\citet{ji2023ddp} &0.82 &53.7 &0.85 &54.6 \\
~\citet{ji2023ddp}+ours &\textbf{0.88} &\textbf{57.2} &\textbf{0.92} &\textbf{56.6} \\ \hline
\multicolumn{5}{c}{Source-Free Domain Adaptation} \\ \hline
CMA~\cite{bruggemann2023contrastive} &0.79 &50.1 &0.80 &51.3 \\ \hline
~\citet{ji2023ddp} &0.82 &53.7 &0.85 &54.6 \\
~\citet{ji2023ddp}+ours &\textbf{0.86} &\textbf{55.2} &\textbf{0.89} &\textbf{55.9} \\
\bottomrule
\end{tabular}}
\end{table}
\vspace{-5pt}

\subsection{Semantic Segmentation}
\noindent\textbf{Implementation and Datasets} We adopt ~\citet{ji2023ddp} as the baseline. We choose aAcc and mIOU as metrics for evaluating the performance. For inference, we set the timestep to 10. ~\citet{ji2023ddp} adopts CityScapes~\cite{cordts2016cityscapes} for training. Following the setting of previous works, we adopt Dark-Zurich~\cite{sakaridis2020map}, a nighttime dataset for semantic segmentation, and ACDC~\cite{sakaridis2021acdc} for inference.

\noindent\textbf{Performance.} We present the results of ~\citet{ji2023ddp} with our methods compared with previous UDA models~\cite{tranheden2021dacs,hoyer2022daformer,bruggemann2023contrastive} under different DA settings in Tab.~\ref{tab:exp_seg}. On both datasets, our method performs comparably to previous UDA models.
\begin{table}[htbp]
\centering
\caption{Ablation studies of different designs}
\label{tab:ablation}
\resizebox{0.45\textwidth}{!}{
\begin{tabular}{lccc}
\toprule
\multirow{2}{*}{Method} & Marigold & DiffBIR & DDVM\\
% \cmidrule(lr){2} \cmidrule(lr){3} \cmidrule(lr){4}
 & AbsRel $\downarrow$ & psnr$\uparrow$ & EPE$\downarrow$ \\ \hline
\multicolumn{4}{c}{Source-Available Domain Adaptation} \\ \hline
direct alignment &25.5  &23.37  & 9.41  \\
alignment with $\lambda_t$ &\textbf{17.0} &\textbf{25.19} &\textbf{4.93} \\ \hline
\multicolumn{4}{c}{Source-Free Domain Adaptation} \\ \hline
only alignment &21.4  &24.33  &7.24 \\
+linear mask schedule &18.5  &24.57  &6.45   \\
+consistency scaling &20.7  &24.39  &7.12  \\
ours &\textbf{18.2} &\textbf{24.70} &\textbf{6.33}\\
\bottomrule
\end{tabular}}
\end{table}
\vspace{-10pt}

\subsection{Ablation Studies}
For Domain Noise Alignment under source-available DA setting, we conduct ablation experiments on the method of noise calculation, comparing our approach with directly scaling the noise to match the noise predictions from the source domain, as shown in~Tab.~\ref{tab:ablation}.  Directly scaling the noise significantly degrades the model's performance, as it severely impacts the generative ability of diffusion models.

For Source-Free Domain Noise Alignment, we performed ablation studies on our three additional design components. It can be observed that both the linear threshold schedule and noise consistency scaling contribute to improved model performance on the target domain.

\section{Conclusion}
\label{sec:conclusion}
In this work, we first analyze the domain bias within image-to-image diffusion models when conditional images originate from disparate domains, starting from the exposure bias inherent in diffusion models. To alleviate the domain bias and enhance the performance of diffusion models on the target domain, we introduce a domain adaptation method specifically tailored for image-to-image translation diffusion models. Moreover, we have extended our consideration to scenarios where source domain data is unavailable. We hope this work will draw increased attention to the generalization challenges faced by diffusion models.

{
    \small
    \bibliographystyle{ieeenat_fullname}
    \bibliography{main}

\begin{thebibliography}{86}
\providecommand{\natexlab}[1]{#1}
\providecommand{\url}[1]{\texttt{#1}}
\expandafter\ifx\csname urlstyle\endcsname\relax
  \providecommand{\doi}[1]{doi: #1}\else
  \providecommand{\doi}{doi: \begingroup \urlstyle{rm}\Url}\fi

\bibitem[Agustsson and Timofte(2017)]{agustsson2017ntire}
Eirikur Agustsson and Radu Timofte.
\newblock Ntire 2017 challenge on single image super-resolution: Dataset and
  study.
\newblock In \emph{Proceedings of the IEEE conference on computer vision and
  pattern recognition workshops}, pages 126--135, 2017.

\bibitem[Amit et~al.(2021)Amit, Shaharbany, Nachmani, and
  Wolf]{amit2021segdiff}
Tomer Amit, Tal Shaharbany, Eliya Nachmani, and Lior Wolf.
\newblock Segdiff: Image segmentation with diffusion probabilistic models.
\newblock \emph{arXiv preprint arXiv:2112.00390}, 2021.

\bibitem[Ayyoubzadeh et~al.(2023)Ayyoubzadeh, Liu, Kezele, Yu, Wu, Wang, and
  Jin]{ayyoubzadeh2023test}
Seyed~Mehdi Ayyoubzadeh, Wentao Liu, Irina Kezele, Yuanhao Yu, Xiaolin Wu, Yang
  Wang, and Tang Jin.
\newblock Test-time adaptation for optical flow estimation using motion
  vectors.
\newblock \emph{IEEE Transactions on Image Processing}, 32:\penalty0
  4977--4988, 2023.

\bibitem[Bao et~al.(2022)Bao, Li, Zhu, and Zhang]{bao2022analytic}
Fan Bao, Chongxuan Li, Jun Zhu, and Bo Zhang.
\newblock Analytic-dpm: an analytic estimate of the optimal reverse variance in
  diffusion probabilistic models.
\newblock \emph{arXiv preprint arXiv:2201.06503}, 2022.

\bibitem[Ben-David et~al.(2006)Ben-David, Blitzer, Crammer, and
  Pereira]{ben2006analysis}
Shai Ben-David, John Blitzer, Koby Crammer, and Fernando Pereira.
\newblock Analysis of representations for domain adaptation.
\newblock \emph{Advances in neural information processing systems}, 19, 2006.

\bibitem[Benigmim et~al.(2023)Benigmim, Roy, Essid, Kalogeiton, and
  Lathuili{\`e}re]{benigmim2023one}
Yasser Benigmim, Subhankar Roy, Slim Essid, Vicky Kalogeiton, and St{\'e}phane
  Lathuili{\`e}re.
\newblock One-shot unsupervised domain adaptation with personalized diffusion
  models.
\newblock In \emph{Proceedings of the IEEE/CVF conference on computer vision
  and pattern recognition}, pages 698--708, 2023.

\bibitem[Br{\"u}ggemann et~al.(2023)Br{\"u}ggemann, Sakaridis, Br{\"o}dermann,
  and Van~Gool]{bruggemann2023contrastive}
David Br{\"u}ggemann, Christos Sakaridis, Tim Br{\"o}dermann, and Luc Van~Gool.
\newblock Contrastive model adaptation for cross-condition robustness in
  semantic segmentation.
\newblock In \emph{Proceedings of the IEEE/CVF International Conference on
  Computer Vision}, pages 11378--11387, 2023.

\bibitem[Butler et~al.(2012)Butler, Wulff, Stanley, and
  Black]{butler2012naturalistic}
Daniel~J Butler, Jonas Wulff, Garrett~B Stanley, and Michael~J Black.
\newblock A naturalistic open source movie for optical flow evaluation.
\newblock In \emph{Computer Vision--ECCV 2012: 12th European Conference on
  Computer Vision, Florence, Italy, October 7-13, 2012, Proceedings, Part VI
  12}, pages 611--625. Springer, 2012.

\bibitem[Caesar et~al.(2020)Caesar, Bankiti, Lang, Vora, Liong, Xu, Krishnan,
  Pan, Baldan, and Beijbom]{caesar2020nuscenes}
Holger Caesar, Varun Bankiti, Alex~H Lang, Sourabh Vora, Venice~Erin Liong,
  Qiang Xu, Anush Krishnan, Yu Pan, Giancarlo Baldan, and Oscar Beijbom.
\newblock nuscenes: A multimodal dataset for autonomous driving.
\newblock In \emph{Proceedings of the IEEE/CVF conference on computer vision
  and pattern recognition}, pages 11621--11631, 2020.

\bibitem[Cai et~al.(2019)Cai, Zeng, Yong, Cao, and Zhang]{cai2019toward}
Jianrui Cai, Hui Zeng, Hongwei Yong, Zisheng Cao, and Lei Zhang.
\newblock Toward real-world single image super-resolution: A new benchmark and
  a new model.
\newblock In \emph{Proceedings of the IEEE/CVF international conference on
  computer vision}, pages 3086--3095, 2019.

\bibitem[Chen et~al.(2019)Chen, Xie, Huang, Rong, Ding, Huang, Xu, and
  Huang]{chen2019progressive}
Chaoqi Chen, Weiping Xie, Wenbing Huang, Yu Rong, Xinghao Ding, Yue Huang,
  Tingyang Xu, and Junzhou Huang.
\newblock Progressive feature alignment for unsupervised domain adaptation.
\newblock In \emph{Proceedings of the IEEE/CVF conference on computer vision
  and pattern recognition}, pages 627--636, 2019.

\bibitem[Chen et~al.(2021{\natexlab{a}})Chen, Peng, Ma, Li, Du, and
  Tian]{chen2021amplitude}
Guangyao Chen, Peixi Peng, Li Ma, Jia Li, Lin Du, and Yonghong Tian.
\newblock Amplitude-phase recombination: Rethinking robustness of convolutional
  neural networks in frequency domain.
\newblock In \emph{Proceedings of the IEEE/CVF international conference on
  computer vision}, pages 458--467, 2021{\natexlab{a}}.

\bibitem[Chen et~al.(2021{\natexlab{b}})Chen, Wang, Chen, and
  Zeng]{chen2021s2r}
Xiaotian Chen, Yuwang Wang, Xuejin Chen, and Wenjun Zeng.
\newblock S2r-depthnet: Learning a generalizable depth-specific structural
  representation.
\newblock In \emph{Proceedings of the IEEE/CVF conference on computer vision
  and pattern recognition}, pages 3034--3043, 2021{\natexlab{b}}.

\bibitem[Clark and Jaini(2023)]{clark2023text}
Kevin Clark and Priyank Jaini.
\newblock Text-to-image diffusion models are zero shot classifiers.
\newblock \emph{Advances in Neural Information Processing Systems},
  36:\penalty0 58921--58937, 2023.

\bibitem[Cordts et~al.(2016)Cordts, Omran, Ramos, Rehfeld, Enzweiler, Benenson,
  Franke, Roth, and Schiele]{cordts2016cityscapes}
Marius Cordts, Mohamed Omran, Sebastian Ramos, Timo Rehfeld, Markus Enzweiler,
  Rodrigo Benenson, Uwe Franke, Stefan Roth, and Bernt Schiele.
\newblock The cityscapes dataset for semantic urban scene understanding.
\newblock In \emph{Proceedings of the IEEE conference on computer vision and
  pattern recognition}, pages 3213--3223, 2016.

\bibitem[Courty et~al.(2017)Courty, Flamary, Habrard, and
  Rakotomamonjy]{courty2017joint}
Nicolas Courty, R{\'e}mi Flamary, Amaury Habrard, and Alain Rakotomamonjy.
\newblock Joint distribution optimal transportation for domain adaptation.
\newblock \emph{Advances in neural information processing systems}, 30, 2017.

\bibitem[Csurka(2017)]{csurka2017domain}
Gabriela Csurka.
\newblock Domain adaptation for visual applications: A comprehensive survey.
\newblock \emph{arXiv preprint arXiv:1702.05374}, 2017.

\bibitem[Dong et~al.(2023)Dong, Zhao, and Fu]{dong2023open}
Qiaole Dong, Bo Zhao, and Yanwei Fu.
\newblock Open-ddvm: A reproduction and extension of diffusion model for
  optical flow estimation.
\newblock \emph{arXiv preprint arXiv:2312.01746}, 2023.

\bibitem[Dziugaite et~al.(2015)Dziugaite, Roy, and
  Ghahramani]{dziugaite2015training}
Gintare~Karolina Dziugaite, Daniel~M Roy, and Zoubin Ghahramani.
\newblock Training generative neural networks via maximum mean discrepancy
  optimization.
\newblock \emph{arXiv preprint arXiv:1505.03906}, 2015.

\bibitem[Farahani et~al.(2021)Farahani, Voghoei, Rasheed, and
  Arabnia]{farahani2021brief}
Abolfazl Farahani, Sahar Voghoei, Khaled Rasheed, and Hamid~R Arabnia.
\newblock A brief review of domain adaptation.
\newblock \emph{Advances in data science and information engineering:
  proceedings from ICDATA 2020 and IKE 2020}, pages 877--894, 2021.

\bibitem[Gaidon et~al.(2016)Gaidon, Wang, Cabon, and Vig]{gaidon2016virtual}
Adrien Gaidon, Qiao Wang, Yohann Cabon, and Eleonora Vig.
\newblock Virtual worlds as proxy for multi-object tracking analysis.
\newblock In \emph{Proceedings of the IEEE conference on computer vision and
  pattern recognition}, pages 4340--4349, 2016.

\bibitem[Ganin et~al.(2016)Ganin, Ustinova, Ajakan, Germain, Larochelle,
  Laviolette, March, and Lempitsky]{ganin2016domain}
Yaroslav Ganin, Evgeniya Ustinova, Hana Ajakan, Pascal Germain, Hugo
  Larochelle, Fran{\c{c}}ois Laviolette, Mario March, and Victor Lempitsky.
\newblock Domain-adversarial training of neural networks.
\newblock \emph{Journal of machine learning research}, 17\penalty0
  (59):\penalty0 1--35, 2016.

\bibitem[He et~al.(2024)He, Li, Yin, Liang, Li, Zhou, Zhang, Liu, and
  Chen]{he2024lotus}
Jing He, Haodong Li, Wei Yin, Yixun Liang, Leheng Li, Kaiqiang Zhou, Hongbo
  Zhang, Bingbing Liu, and Ying-Cong Chen.
\newblock Lotus: Diffusion-based visual foundation model for high-quality dense
  prediction.
\newblock \emph{arXiv preprint arXiv:2409.18124}, 2024.

\bibitem[Hemati et~al.(2023)Hemati, Beitollahi, Estiri, Omari, Chen, and
  Zhang]{hemati2023cross}
Sobhan Hemati, Mahdi Beitollahi, Amir~Hossein Estiri, Bassel~Al Omari, Xi Chen,
  and Guojun Zhang.
\newblock Cross domain generative augmentation: domain generalization with
  latent diffusion models.
\newblock \emph{arXiv preprint arXiv:2312.05387}, 2023.

\bibitem[Ho and Salimans(2022)]{ho2022classifier}
Jonathan Ho and Tim Salimans.
\newblock Classifier-free diffusion guidance.
\newblock \emph{arXiv preprint arXiv:2207.12598}, 2022.

\bibitem[Ho et~al.(2022)Ho, Saharia, Chan, Fleet, Norouzi, and
  Salimans]{ho2022cascaded}
Jonathan Ho, Chitwan Saharia, William Chan, David~J Fleet, Mohammad Norouzi,
  and Tim Salimans.
\newblock Cascaded diffusion models for high fidelity image generation.
\newblock \emph{Journal of Machine Learning Research}, 23\penalty0
  (47):\penalty0 1--33, 2022.

\bibitem[Hoyer et~al.(2022)Hoyer, Dai, and Van~Gool]{hoyer2022daformer}
Lukas Hoyer, Dengxin Dai, and Luc Van~Gool.
\newblock Daformer: Improving network architectures and training strategies for
  domain-adaptive semantic segmentation.
\newblock In \emph{Proceedings of the IEEE/CVF conference on computer vision
  and pattern recognition}, pages 9924--9935, 2022.

\bibitem[Huang and Belongie(2017)]{huang2017arbitrary}
Xun Huang and Serge Belongie.
\newblock Arbitrary style transfer in real-time with adaptive instance
  normalization.
\newblock In \emph{Proceedings of the IEEE international conference on computer
  vision}, pages 1501--1510, 2017.

\bibitem[Ji et~al.(2024)Ji, Lin, and Li]{ji2024dpbridge}
Haorui Ji, Taojun Lin, and Hongdong Li.
\newblock Dpbridge: Latent diffusion bridge for dense prediction.
\newblock \emph{arXiv preprint arXiv:2412.20506}, 2024.

\bibitem[Ji et~al.(2023)Ji, Chen, Xie, Hong, Liu, Liu, Lu, Li, and
  Luo]{ji2023ddp}
Yuanfeng Ji, Zhe Chen, Enze Xie, Lanqing Hong, Xihui Liu, Zhaoqiang Liu, Tong
  Lu, Zhenguo Li, and Ping Luo.
\newblock Ddp: Diffusion model for dense visual prediction.
\newblock In \emph{Proceedings of the IEEE/CVF International Conference on
  Computer Vision}, pages 21741--21752, 2023.

\bibitem[Ke et~al.(2024)Ke, Obukhov, Huang, Metzger, Daudt, and
  Schindler]{ke2024repurposing}
Bingxin Ke, Anton Obukhov, Shengyu Huang, Nando Metzger, Rodrigo~Caye Daudt,
  and Konrad Schindler.
\newblock Repurposing diffusion-based image generators for monocular depth
  estimation.
\newblock In \emph{Proceedings of the IEEE/CVF Conference on Computer Vision
  and Pattern Recognition}, pages 9492--9502, 2024.

\bibitem[Ke et~al.(2021)Ke, Wang, Wang, Milanfar, and Yang]{ke2021musiq}
Junjie Ke, Qifei Wang, Yilin Wang, Peyman Milanfar, and Feng Yang.
\newblock Musiq: Multi-scale image quality transformer.
\newblock In \emph{Proceedings of the IEEE/CVF international conference on
  computer vision}, pages 5148--5157, 2021.

\bibitem[Kim et~al.(2023)Kim, Li, and Hospedales]{kim2023domain}
Minyoung Kim, Da Li, and Timothy Hospedales.
\newblock Domain generalisation via domain adaptation: An adversarial fourier
  amplitude approach.
\newblock \emph{arXiv preprint arXiv:2302.12047}, 2023.

\bibitem[Kim and Kim(2021)]{kim2021semi}
Yoonhyung Kim and Changick Kim.
\newblock Semi-supervised domain adaptation via selective pseudo labeling and
  progressive self-training.
\newblock In \emph{2020 25th International Conference on Pattern Recognition
  (ICPR)}, pages 1059--1066. IEEE, 2021.

\bibitem[Lee et~al.(2024)Lee, Tseng, and Yang]{lee2024exploiting}
Hsin-Ying Lee, Hung-Yu Tseng, and Ming-Hsuan Yang.
\newblock Exploiting diffusion prior for generalizable dense prediction.
\newblock In \emph{Proceedings of the IEEE/CVF Conference on Computer Vision
  and Pattern Recognition}, pages 7861--7871, 2024.

\bibitem[Li et~al.(2023{\natexlab{a}})Li, Prabhudesai, Duggal, Brown, and
  Pathak]{li2023your}
Alexander~C Li, Mihir Prabhudesai, Shivam Duggal, Ellis Brown, and Deepak
  Pathak.
\newblock Your diffusion model is secretly a zero-shot classifier.
\newblock In \emph{Proceedings of the IEEE/CVF International Conference on
  Computer Vision}, pages 2206--2217, 2023{\natexlab{a}}.

\bibitem[Li et~al.(2020)Li, Chen, Ding, Zhu, Lu, and Shen]{li2020maximum}
Jingjing Li, Erpeng Chen, Zhengming Ding, Lei Zhu, Ke Lu, and Heng~Tao Shen.
\newblock Maximum density divergence for domain adaptation.
\newblock \emph{IEEE transactions on pattern analysis and machine
  intelligence}, 43\penalty0 (11):\penalty0 3918--3930, 2020.

\bibitem[Li et~al.(2023{\natexlab{b}})Li, Qu, Yao, Sun, and
  Moens]{li2023alleviating}
Mingxiao Li, Tingyu Qu, Ruicong Yao, Wei Sun, and Marie-Francine Moens.
\newblock Alleviating exposure bias in diffusion models through sampling with
  shifted time steps.
\newblock \emph{arXiv preprint arXiv:2305.15583}, 2023{\natexlab{b}}.

\bibitem[Li and Chen(2022)]{li2022unsupervised}
Weikai Li and Songcan Chen.
\newblock Unsupervised domain adaptation with progressive adaptation of
  subspaces.
\newblock \emph{Pattern Recognition}, 132:\penalty0 108918, 2022.

\bibitem[Li et~al.(2023{\natexlab{c}})Li, Shi, Schiele, and Dai]{li2023test}
Zhi Li, Shaoshuai Shi, Bernt Schiele, and Dengxin Dai.
\newblock Test-time domain adaptation for monocular depth estimation.
\newblock In \emph{2023 IEEE International Conference on Robotics and
  Automation (ICRA)}, pages 4873--4879. IEEE, 2023{\natexlab{c}}.

\bibitem[Liang et~al.(2021)Liang, Cao, Sun, Zhang, Van~Gool, and
  Timofte]{liang2021swinir}
Jingyun Liang, Jiezhang Cao, Guolei Sun, Kai Zhang, Luc Van~Gool, and Radu
  Timofte.
\newblock Swinir: Image restoration using swin transformer.
\newblock In \emph{Proceedings of the IEEE/CVF international conference on
  computer vision}, pages 1833--1844, 2021.

\bibitem[Lin et~al.(2024{\natexlab{a}})Lin, Liu, Li, and Yang]{lin2024common}
Shanchuan Lin, Bingchen Liu, Jiashi Li, and Xiao Yang.
\newblock Common diffusion noise schedules and sample steps are flawed.
\newblock In \emph{Proceedings of the IEEE/CVF winter conference on
  applications of computer vision}, pages 5404--5411, 2024{\natexlab{a}}.

\bibitem[Lin et~al.(2024{\natexlab{b}})Lin, He, Chen, Lyu, Dai, Yu, Qiao,
  Ouyang, and Dong]{lin2024diffbir}
Xinqi Lin, Jingwen He, Ziyan Chen, Zhaoyang Lyu, Bo Dai, Fanghua Yu, Yu Qiao,
  Wanli Ouyang, and Chao Dong.
\newblock Diffbir: Toward blind image restoration with generative diffusion
  prior.
\newblock In \emph{European Conference on Computer Vision}, pages 430--448.
  Springer, 2024{\natexlab{b}}.

\bibitem[London et~al.(2016)London, Huang, and Getoor]{london2016stability}
Ben London, Bert Huang, and Lise Getoor.
\newblock Stability and generalization in structured prediction.
\newblock \emph{Journal of Machine Learning Research}, 17\penalty0
  (221):\penalty0 1--52, 2016.

\bibitem[Long et~al.(2018)Long, Cao, Wang, and Jordan]{long2018conditional}
Mingsheng Long, Zhangjie Cao, Jianmin Wang, and Michael~I Jordan.
\newblock Conditional adversarial domain adaptation.
\newblock \emph{Advances in neural information processing systems}, 31, 2018.

\bibitem[Lopez-Rodriguez and Mikolajczyk(2023)]{lopez2023desc}
Adrian Lopez-Rodriguez and Krystian Mikolajczyk.
\newblock Desc: Domain adaptation for depth estimation via semantic
  consistency.
\newblock \emph{International Journal of Computer Vision}, 131\penalty0
  (3):\penalty0 752--771, 2023.

\bibitem[Luo et~al.(2024)Luo, Li, Yang, Liu, Fan, and Liu]{luo2024flowdiffuser}
Ao Luo, Xin Li, Fan Yang, Jiangyu Liu, Haoqiang Fan, and Shuaicheng Liu.
\newblock Flowdiffuser: Advancing optical flow estimation with diffusion
  models.
\newblock In \emph{Proceedings of the IEEE/CVF Conference on Computer Vision
  and Pattern Recognition}, pages 19167--19176, 2024.

\bibitem[Maddern et~al.(2017)Maddern, Pascoe, Linegar, and
  Newman]{maddern20171}
Will Maddern, Geoffrey Pascoe, Chris Linegar, and Paul Newman.
\newblock 1 year, 1000 km: The oxford robotcar dataset.
\newblock \emph{The International Journal of Robotics Research}, 36\penalty0
  (1):\penalty0 3--15, 2017.

\bibitem[Niemeijer et~al.(2024)Niemeijer, Schwonberg, Term{\"o}hlen, Schmidt,
  and Fingscheidt]{niemeijer2024generalization}
Joshua Niemeijer, Manuel Schwonberg, Jan-Aike Term{\"o}hlen, Nico~M Schmidt,
  and Tim Fingscheidt.
\newblock Generalization by adaptation: Diffusion-based domain extension for
  domain-generalized semantic segmentation.
\newblock In \emph{Proceedings of the IEEE/CVF Winter Conference on
  Applications of Computer Vision}, pages 2830--2840, 2024.

\bibitem[Ning et~al.(2023{\natexlab{a}})Ning, Li, Su, Salah, and
  Ertugrul]{ning2023elucidating}
Mang Ning, Mingxiao Li, Jianlin Su, Albert~Ali Salah, and Itir~Onal Ertugrul.
\newblock Elucidating the exposure bias in diffusion models.
\newblock \emph{arXiv preprint arXiv:2308.15321}, 2023{\natexlab{a}}.

\bibitem[Ning et~al.(2023{\natexlab{b}})Ning, Sangineto, Porrello, Calderara,
  and Cucchiara]{ning2023input}
Mang Ning, Enver Sangineto, Angelo Porrello, Simone Calderara, and Rita
  Cucchiara.
\newblock Input perturbation reduces exposure bias in diffusion models.
\newblock \emph{arXiv preprint arXiv:2301.11706}, 2023{\natexlab{b}}.

\bibitem[Ranftl et~al.(2020)Ranftl, Lasinger, Hafner, Schindler, and
  Koltun]{ranftl2020towards}
Ren{\'e} Ranftl, Katrin Lasinger, David Hafner, Konrad Schindler, and Vladlen
  Koltun.
\newblock Towards robust monocular depth estimation: Mixing datasets for
  zero-shot cross-dataset transfer.
\newblock \emph{IEEE transactions on pattern analysis and machine
  intelligence}, 44\penalty0 (3):\penalty0 1623--1637, 2020.

\bibitem[Ranftl et~al.(2021)Ranftl, Bochkovskiy, and Koltun]{ranftl2021vision}
Ren{\'e} Ranftl, Alexey Bochkovskiy, and Vladlen Koltun.
\newblock Vision transformers for dense prediction.
\newblock In \emph{Proceedings of the IEEE/CVF international conference on
  computer vision}, pages 12179--12188, 2021.

\bibitem[Ren et~al.(2024)Ren, Zhan, Ding, Wang, Wang, Fan, and
  Tao]{ren2024multi}
Zhiyao Ren, Yibing Zhan, Liang Ding, Gaoang Wang, Chaoyue Wang, Zhongyi Fan,
  and Dacheng Tao.
\newblock Multi-step denoising scheduled sampling: Towards alleviating exposure
  bias for diffusion models.
\newblock In \emph{Proceedings of the AAAI Conference on Artificial
  Intelligence}, pages 4667--4675, 2024.

\bibitem[Roberts et~al.(2021)Roberts, Ramapuram, Ranjan, Kumar, Bautista,
  Paczan, Webb, and Susskind]{roberts2021hypersim}
Mike Roberts, Jason Ramapuram, Anurag Ranjan, Atulit Kumar, Miguel~Angel
  Bautista, Nathan Paczan, Russ Webb, and Joshua~M Susskind.
\newblock Hypersim: A photorealistic synthetic dataset for holistic indoor
  scene understanding.
\newblock In \emph{Proceedings of the IEEE/CVF international conference on
  computer vision}, pages 10912--10922, 2021.

\bibitem[Rombach et~al.(2022)Rombach, Blattmann, Lorenz, Esser, and
  Ommer]{rombach2022high}
Robin Rombach, Andreas Blattmann, Dominik Lorenz, Patrick Esser, and Bj{\"o}rn
  Ommer.
\newblock High-resolution image synthesis with latent diffusion models.
\newblock In \emph{Proceedings of the IEEE/CVF conference on computer vision
  and pattern recognition}, pages 10684--10695, 2022.

\bibitem[Saito et~al.(2020)Saito, Kim, Sclaroff, and
  Saenko]{saito2020universal}
Kuniaki Saito, Donghyun Kim, Stan Sclaroff, and Kate Saenko.
\newblock Universal domain adaptation through self supervision.
\newblock \emph{Advances in neural information processing systems},
  33:\penalty0 16282--16292, 2020.

\bibitem[Sakaridis et~al.(2020)Sakaridis, Dai, and Van~Gool]{sakaridis2020map}
Christos Sakaridis, Dengxin Dai, and Luc Van~Gool.
\newblock Map-guided curriculum domain adaptation and uncertainty-aware
  evaluation for semantic nighttime image segmentation.
\newblock \emph{IEEE Transactions on Pattern Analysis and Machine
  Intelligence}, 44\penalty0 (6):\penalty0 3139--3153, 2020.

\bibitem[Sakaridis et~al.(2021)Sakaridis, Dai, and Van~Gool]{sakaridis2021acdc}
Christos Sakaridis, Dengxin Dai, and Luc Van~Gool.
\newblock Acdc: The adverse conditions dataset with correspondences for
  semantic driving scene understanding.
\newblock In \emph{Proceedings of the IEEE/CVF international conference on
  computer vision}, pages 10765--10775, 2021.

\bibitem[Saxena et~al.(2023)Saxena, Herrmann, Hur, Kar, Norouzi, Sun, and
  Fleet]{saxena2023surprising}
Saurabh Saxena, Charles Herrmann, Junhwa Hur, Abhishek Kar, Mohammad Norouzi,
  Deqing Sun, and David~J Fleet.
\newblock The surprising effectiveness of diffusion models for optical flow and
  monocular depth estimation.
\newblock \emph{Advances in Neural Information Processing Systems},
  36:\penalty0 39443--39469, 2023.

\bibitem[Schuhmann et~al.(2022)Schuhmann, Beaumont, Vencu, Gordon, Wightman,
  Cherti, Coombes, Katta, Mullis, Wortsman, et~al.]{schuhmann2022laion}
Christoph Schuhmann, Romain Beaumont, Richard Vencu, Cade Gordon, Ross
  Wightman, Mehdi Cherti, Theo Coombes, Aarush Katta, Clayton Mullis, Mitchell
  Wortsman, et~al.
\newblock Laion-5b: An open large-scale dataset for training next generation
  image-text models.
\newblock \emph{Advances in neural information processing systems},
  35:\penalty0 25278--25294, 2022.

\bibitem[Silberman et~al.(2012)Silberman, Hoiem, Kohli, and
  Fergus]{silberman2012indoor}
Nathan Silberman, Derek Hoiem, Pushmeet Kohli, and Rob Fergus.
\newblock Indoor segmentation and support inference from rgbd images.
\newblock In \emph{Computer Vision--ECCV 2012: 12th European Conference on
  Computer Vision, Florence, Italy, October 7-13, 2012, Proceedings, Part V
  12}, pages 746--760. Springer, 2012.

\bibitem[Song et~al.(2022)Song, Han, Liu, Metaxas, and
  Elgammal]{song2022diffusion}
Kunpeng Song, Ligong Han, Bingchen Liu, Dimitris Metaxas, and Ahmed Elgammal.
\newblock Diffusion guided domain adaptation of image generators.
\newblock \emph{arXiv preprint arXiv:2212.04473}, 2022.

\bibitem[Sun et~al.(2021)Sun, Vlasic, Herrmann, Jampani, Krainin, Chang, Zabih,
  Freeman, and Liu]{sun2021autoflow}
Deqing Sun, Daniel Vlasic, Charles Herrmann, Varun Jampani, Michael Krainin,
  Huiwen Chang, Ramin Zabih, William~T Freeman, and Ce Liu.
\newblock Autoflow: Learning a better training set for optical flow.
\newblock In \emph{Proceedings of the IEEE/CVF Conference on Computer Vision
  and Pattern Recognition}, pages 10093--10102, 2021.

\bibitem[Sun et~al.(2019)Sun, Tzeng, Darrell, and Efros]{sun2019unsupervised}
Yu Sun, Eric Tzeng, Trevor Darrell, and Alexei~A Efros.
\newblock Unsupervised domain adaptation through self-supervision.
\newblock \emph{arXiv preprint arXiv:1909.11825}, 2019.

\bibitem[Tateno et~al.(2018)Tateno, Navab, and Tombari]{tateno2018distortion}
Keisuke Tateno, Nassir Navab, and Federico Tombari.
\newblock Distortion-aware convolutional filters for dense prediction in
  panoramic images.
\newblock In \emph{Proceedings of the European Conference on Computer Vision
  (ECCV)}, pages 707--722, 2018.

\bibitem[Timofte et~al.(2017)Timofte, Agustsson, Van~Gool, Yang, and
  Zhang]{timofte2017ntire}
Radu Timofte, Eirikur Agustsson, Luc Van~Gool, Ming-Hsuan Yang, and Lei Zhang.
\newblock Ntire 2017 challenge on single image super-resolution: Methods and
  results.
\newblock In \emph{Proceedings of the IEEE conference on computer vision and
  pattern recognition workshops}, pages 114--125, 2017.

\bibitem[Tranheden et~al.(2021)Tranheden, Olsson, Pinto, and
  Svensson]{tranheden2021dacs}
Wilhelm Tranheden, Viktor Olsson, Juliano Pinto, and Lennart Svensson.
\newblock Dacs: Domain adaptation via cross-domain mixed sampling.
\newblock In \emph{Proceedings of the IEEE/CVF winter conference on
  applications of computer vision}, pages 1379--1389, 2021.

\bibitem[Tzeng et~al.(2017)Tzeng, Hoffman, Saenko, and
  Darrell]{tzeng2017adversarial}
Eric Tzeng, Judy Hoffman, Kate Saenko, and Trevor Darrell.
\newblock Adversarial discriminative domain adaptation.
\newblock In \emph{Proceedings of the IEEE conference on computer vision and
  pattern recognition}, pages 7167--7176, 2017.

\bibitem[Wang et~al.(2022)Wang, Fan, Cai, Liu, and Wang]{wang2022undaf}
Hengli Wang, Rui Fan, Peide Cai, Ming Liu, and Lujia Wang.
\newblock Undaf: A general unsupervised domain adaptation framework for
  disparity or optical flow estimation.
\newblock In \emph{2022 International Conference on Robotics and Automation
  (ICRA)}, pages 01--07. IEEE, 2022.

\bibitem[Wang et~al.(2024)Wang, Yue, Zhou, Chan, and Loy]{wang2024exploiting}
Jianyi Wang, Zongsheng Yue, Shangchen Zhou, Kelvin~CK Chan, and Chen~Change
  Loy.
\newblock Exploiting diffusion prior for real-world image super-resolution.
\newblock \emph{International Journal of Computer Vision}, 132\penalty0
  (12):\penalty0 5929--5949, 2024.

\bibitem[Wang and Deng(2018)]{wang2018deep}
Mei Wang and Weihong Deng.
\newblock Deep visual domain adaptation: A survey.
\newblock \emph{Neurocomputing}, 312:\penalty0 135--153, 2018.

\bibitem[Wang et~al.(2021{\natexlab{a}})Wang, Li, Ding, Nie, Chen, Dong, and
  Wang]{wang2021rethinking}
Wei Wang, Haojie Li, Zhengming Ding, Feiping Nie, Junyang Chen, Xiao Dong, and
  Zhihui Wang.
\newblock Rethinking maximum mean discrepancy for visual domain adaptation.
\newblock \emph{IEEE Transactions on Neural Networks and Learning Systems},
  34\penalty0 (1):\penalty0 264--277, 2021{\natexlab{a}}.

\bibitem[Wang et~al.(2021{\natexlab{b}})Wang, Xie, Li, Fan, Song, Liang, Lu,
  Luo, and Shao]{wang2021pyramid}
Wenhai Wang, Enze Xie, Xiang Li, Deng-Ping Fan, Kaitao Song, Ding Liang, Tong
  Lu, Ping Luo, and Ling Shao.
\newblock Pyramid vision transformer: A versatile backbone for dense prediction
  without convolutions.
\newblock In \emph{Proceedings of the IEEE/CVF international conference on
  computer vision}, pages 568--578, 2021{\natexlab{b}}.

\bibitem[Wei et~al.(2020)Wei, Xie, Lu, Zhan, Ye, Zuo, and
  Lin]{wei2020component}
Pengxu Wei, Ziwei Xie, Hannan Lu, Zongyuan Zhan, Qixiang Ye, Wangmeng Zuo, and
  Liang Lin.
\newblock Component divide-and-conquer for real-world image super-resolution.
\newblock In \emph{Computer Vision--ECCV 2020: 16th European Conference,
  Glasgow, UK, August 23--28, 2020, Proceedings, Part VIII 16}, pages 101--117.
  Springer, 2020.

\bibitem[Xiong et~al.(2021)Xiong, Tang, Deng, Zhao, and Yu]{xiong2021multi}
Peng Xiong, Baoping Tang, Lei Deng, Minghang Zhao, and Xiaoxia Yu.
\newblock Multi-block domain adaptation with central moment discrepancy for
  fault diagnosis.
\newblock \emph{Measurement}, 169:\penalty0 108516, 2021.

\bibitem[Yang et~al.(2022)Yang, Wu, Shi, Lao, Gong, Cao, Wang, and
  Yang]{yang2022maniqa}
Sidi Yang, Tianhe Wu, Shuwei Shi, Shanshan Lao, Yuan Gong, Mingdeng Cao, Jiahao
  Wang, and Yujiu Yang.
\newblock Maniqa: Multi-dimension attention network for no-reference image
  quality assessment.
\newblock In \emph{Proceedings of the IEEE/CVF conference on computer vision
  and pattern recognition}, pages 1191--1200, 2022.

\bibitem[Yin et~al.(2021)Yin, Zhang, Wang, Niklaus, Mai, Chen, and Shen]{LeReS}
Wei Yin, Jianming Zhang, Oliver Wang, Simon Niklaus, Long Mai, Simon Chen, and
  Chunhua Shen.
\newblock Learning to recover 3d scene shape from a single image.
\newblock In \emph{Proceedings of the IEEE/CVF Conference on Computer Vision
  and Pattern Recognition}, pages 204--213, 2021.

\bibitem[Yoon et~al.(2024)Yoon, Kim, Kwak, and Cho]{yoon2024optical}
Jeongbeen Yoon, Sanghyun Kim, Suha Kwak, and Minsu Cho.
\newblock Optical flow domain adaptation via target style transfer.
\newblock In \emph{Proceedings of the IEEE/CVF Winter Conference on
  Applications of Computer Vision}, pages 2111--2121, 2024.

\bibitem[Yuan et~al.(2021)Yuan, Fu, Huang, Lin, Zhang, Chen, and
  Wang]{yuan2021hrformer}
Yuhui Yuan, Rao Fu, Lang Huang, Weihong Lin, Chao Zhang, Xilin Chen, and
  Jingdong Wang.
\newblock Hrformer: High-resolution transformer for dense prediction.
\newblock \emph{arXiv preprint arXiv:2110.09408}, 2021.

\bibitem[Zellinger et~al.(2017)Zellinger, Grubinger, Lughofer, Natschl{\"a}ger,
  and Saminger-Platz]{zellinger2017central}
Werner Zellinger, Thomas Grubinger, Edwin Lughofer, Thomas Natschl{\"a}ger, and
  Susanne Saminger-Platz.
\newblock Central moment discrepancy (cmd) for domain-invariant representation
  learning.
\newblock \emph{arXiv preprint arXiv:1702.08811}, 2017.

\bibitem[Zhang et~al.(2024)Zhang, Song, Shi, Liu, and Li]{zhang2024three}
Manyuan Zhang, Guanglu Song, Xiaoyu Shi, Yu Liu, and Hongsheng Li.
\newblock Three things we need to know about transferring stable diffusion to
  visual dense prediction tasks.
\newblock In \emph{European Conference on Computer Vision}, pages 128--145.
  Springer, 2024.

\bibitem[Zheng et~al.(2020)Zheng, Zhang, and Lu]{zheng2020optical}
Yinqiang Zheng, Mingfang Zhang, and Feng Lu.
\newblock Optical flow in the dark.
\newblock In \emph{Proceedings of the IEEE/CVF conference on computer vision
  and pattern recognition}, pages 6749--6757, 2020.

\bibitem[Zhou et~al.(2023)Zhou, Chang, Yan, and Yan]{zhou2023unsupervised}
Hanyu Zhou, Yi Chang, Wending Yan, and Luxin Yan.
\newblock Unsupervised cumulative domain adaptation for foggy scene optical
  flow.
\newblock In \emph{Proceedings of the IEEE/CVF conference on computer vision
  and pattern recognition}, pages 9569--9578, 2023.

\bibitem[Zhu et~al.(2023)Zhu, Li, Wang, He, and Yao]{zhu2023conditional}
Yuanzhi Zhu, Zhaohai Li, Tianwei Wang, Mengchao He, and Cong Yao.
\newblock Conditional text image generation with diffusion models.
\newblock In \emph{Proceedings of the IEEE/CVF Conference on Computer Vision
  and Pattern Recognition}, pages 14235--14245, 2023.

\bibitem[Zhu et~al.(2024)Zhu, Zhao, Li, Tang, Zhou, and Lu]{zhu2024flowie}
Yixuan Zhu, Wenliang Zhao, Ao Li, Yansong Tang, Jie Zhou, and Jiwen Lu.
\newblock Flowie: Efficient image enhancement via rectified flow.
\newblock In \emph{Proceedings of the IEEE/CVF Conference on Computer Vision
  and Pattern Recognition}, pages 13--22, 2024.

\end{thebibliography}
}

\end{document}